\documentclass{article} 
\usepackage{PRIMEarxiv}

\usepackage{hyperref}
\usepackage{url}

\usepackage{microtype}
\usepackage{graphicx}
\usepackage{subfigure}
\usepackage{booktabs} 
\usepackage{subcaption,wrapfig}
\usepackage{amsmath}
\usepackage{amssymb}
\usepackage{mathtools}
\usepackage{amsthm}
\theoremstyle{plain}
\newtheorem{theorem}{Theorem}[section]

\theoremstyle{definition}

\theoremstyle{remark}

\title{On robust overfitting: adversarial training induced distribution matters}

\author{Runzhi Tian\\
Department of EECS\\
University of Ottawa\\
Ottawa, ON, Canada \\
\texttt{\{rtian081\}@uottawa.ca} \\
\And
Yongyi Mao\\
Department of EECS\\
University of Ottawa\\
Ottawa, ON, Canada \\
\texttt{\{ymao\}@uottawa.ca} \\
}

\begin{document}

\maketitle

\begin{abstract}

Adversarial training may be regarded as standard training with a modified loss function. But its generalization error appears much larger than standard training under standard loss. This phenomenon, known as robust overfitting, has attracted significant research attention  and remains largely as a mystery. In this paper, we first show empirically that robust overfitting correlates with the increasing generalization difficulty of the perturbation-induced distributions along the trajectory of adversarial training (specifically PGD-based adversarial training). We then provide a novel upper bound for generalization error with respect to the perturbation-induced distributions, in which a notion of the perturbation operator, referred to ``local dispersion'',  plays an important role. Experimental results are presented to validate the usefulness of the bound and various additional insights are provided.

\end{abstract}

\section{Introduction}
Despite their outstanding performance, deep neural networks (DNNs) are known to be vulnerable to adversarial attacks where a carefully designed perturbation may cause the network to make a wrong prediction \cite{szegedy2014intriguing, goodfellow2015explaining}. Many methods have been proposed to improve the robustness of DNNs against adversarial perturbations \cite{madry2019deep, DBLP:journals/corr/abs-1901-08573, DBLP:journals/corr/abs-2010-09670}, among which Projected Gradient Descend based Adversarial Training (PGD-AT) \cite{madry2019deep} is arguably the most effective \cite{DBLP:journals/corr/abs-1802-00420, dong2020benchmarking}. A recent work in \cite{DBLP:journals/corr/abs-2002-11569} however revealed a surprising phenomenon in PGD-AT: after training, even though the robust error (i.e., error probability in the predicted label for adversarially perturbed instances) is nearly zero on the training set, it may remain very high on the testing set. For example, on the testing set of CIFAR10 \cite{krizhevsky2009learning}, the robust error of PGD-AT trained model can be as large as 44.19\%. This significantly contrasts the standard training: on CIFAR10, when the standard error (i.e., the error probability in the predicted label for non-perturbed instances) is nearly zero on the training set, its value on the testing set is only about 4\%. This unexpected phenomenon arising in PGD-AT is often referred to as robust overfitting. 
\begin{figure}[!htpb]
    \centering
    \includegraphics[width=0.32\textwidth]{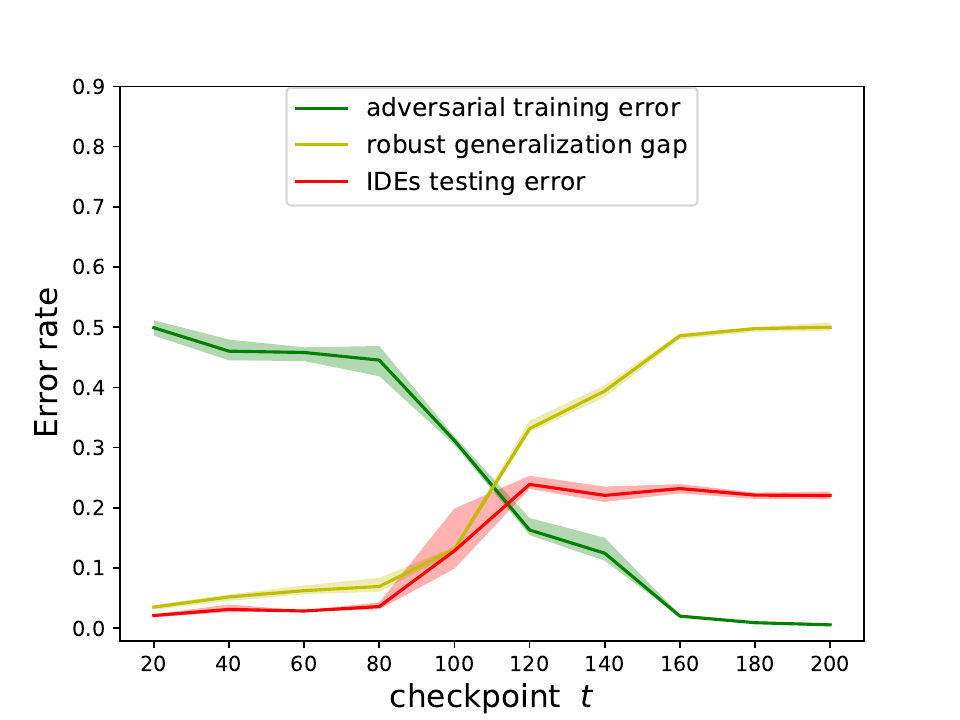}
    \caption{PGD-AT and the corresponding IDE results  on the CIFAR-10 dataset with the Wide ResNet34 (WRN-34) model \cite{DBLP:journals/corr/ZagoruykoK16}. A significant increase in the IDE testing error is observed with the appearance of robust overfitting, suggesting a correlation between the generalization difficulty on $\tilde{\mathcal{D}}_t$ and robust overfitting.}
    \label{fig:IDE cifar10}
\end{figure}

Since its discovery, robust overfitting has attracted significant research attention. A great deal of research effort has been spent on understanding its cause and devising mitigating techniques. The work of \cite{DBLP:journals/corr/abs-2004-05884} and  \cite{DBLP:journals/corr/abs-2104-04448} correlate robust overfitting with the sharpness of the minima in the loss landscape and a method flattening such minima is presented as a remedy. Built on a similar intuition, heuristics such as smoothing the weights or the logit output of neural networks are proposed in \cite{chen2021robust}.  The work of \cite{DBLP:journals/corr/abs-2102-07861} suggests that the robust overfitting is related to the curvature of the activation functions and that low curvature in the activation function appears to improve robust generalization. In  \cite{DBLP:journals/corr/abs-2110-03135}, the authors observe the existence of label noise in PGD-AT and regard it as a source of robust overfitting phenomenon, where the label noise refers to that after adversarial perturbation, the original label may no longer reflect the semantics of the example perfectly. The work of \cite{DBLP:journals/corr/abs-2106-01606} attributes robust overfitting to a memorization effect and label noise in PGD-AT, and subsequently proposes a mitigation algorithm based on an analysis of memorization.  The authors of \cite{yu2022understanding} observe that in PGD-AT, fitting the training examples with smaller adversarial loss tend to cause robust overfitting and propose a heuristic to remove a fraction of the low-loss example during training.  In \cite{smoothness}, robust overfitting is attributed to the non-smoothness loss used in AT, and the authors propose a smoothing technique as a solution.

Encouraging as these progresses are, 
the current understanding of robust overfitting is still arguably far from being conclusive. For example, as pointed out in \cite{hameed2022boundary}, the explanations in \cite{DBLP:journals/corr/abs-2106-01606} and \cite{yu2022understanding} appear to conflict to each other: the former attributes the robust overfitting to the model fitting the data with large adversarial loss while the latter claims that fitting the the data with small adversarial loss is the source of robust overfitting. Furthermore, the proposed mitigation techniques so far, although have been shown to improve generalization, only reduce the testing robust error by a few percent. This may imply that robust overfitting can be due to a multitude of sources, the full picture remaining obscure.

Our work aims at further understanding robust overfitting, attempting to obtain insights by inspecting the dynamics of PGD-AT, the most popular iterative training algorithm. The inspiration of our experimental design stems from the recognition that along the iterations in PGD-AT, adversarial perturbation effectively induces a new data distribution, say $\tilde{\cal D}_t$, at each training step $t$. This distribution, different from the original data distribution ${\cal D}$, continuously evolves in a fashion that depends on the current model parameter $\theta_t$, which in turn affecting the updating of $\theta_t$. It is then curious whether certain properties of $\tilde{\cal D}_t$ or the causes of such properties may be related to robust overfitting. We then conducted a set of experiments, which we call ``induced distribution experiments'' or IDEs, in which we inspect how well a model trained on samples drawn from $\tilde{\cal D}_t$ (under standard training) generalizes. Specifically, a set of time steps, or ``checkpoints'', are selected; at each checkpoint $t$, a model is {\bf trained from scratch} on samples drawn from $\tilde{\cal D}_t$ under the standard loss. Figure \ref{fig:IDE cifar10} shows one such result on CIFAR-10 (for more results, see Section 4). In the figure, the yellow curve, indicating the generalization gap for robust error, gradually elevates as PGD-AT proceeds; the red curve, indicating the IDE testing error, follows a similar trend. The consistent trends of the two curves suggest a correlation between the generalization behaviour of the learned model along the PGD-AT trajectory and the IDE test errors. Noting that the IDE testing error at checkpoint $t$ indicates the generalization difficulty of the induced distribution $\tilde{\cal D}_t$. Thus this observation hints that the perturbation at checkpoint $t$, turning the original distribution ${\cal D}$ to $\tilde{\cal D}_t$,  may be fundamentally related to this difficulty and to robust overfitting.

To further understand the generalization difficulty of $\tilde{\cal D}_t$ and the impact of perturbation on such difficulty, we derive an upper bound of generalization error for distribution $\tilde{\cal D}_t$.  The bound reveals that a key quantity governing the generalization difficulty is a local ``dispersion property'' of the adversarial perturbation at checkpoint $t$: less dispersive perturbations provide better generalization guarantees. This result is corroborated by further empirical observations: when robust overfitting occurs, the adversarial perturbations become increasingly dispersive along the PGD-AT trajectory. 

We also conducted additional experiments to examine the local dispersion of the perturbations along PGD-AT steps. Interestingly, as PGD-AT proceeds, although the perturbation decreases it magnitudes, its directions appear to further spread out. This observation, clearly correlating with the local dispersivenss of adversarial perturbation, is, to our best knowledge, reported for the first time.  It may open new directions in furthering the understanding of robust overfitting.

This paper opens a new dimension in the study of robust overfitting. Not only have we discovered the important roles of the evolving dispersiveness of adversarial perturbation in robust overfitting, this work has also made us believe that, for in-depth understanding of robust overfitting, the dynamics of adversarial training can not be factored out.

\section{Other Related works}
Our work lies in the topic of robust generalization. Different from the standard setting, the robust generalization in deep learning, especially on high dimensional data, seems significantly difficult. Various work have attempted to understand the reason behind. The work in \cite{DBLP:journals/corr/abs-1804-11285} proves that on simple data models such as the Gaussian and Bernoulli models, the robust generalization can be much harder compared to the standard generalization in the sense of sample complexity. The sample complexity of robust generalization is further explored by using conventional statistical learning tools, such as Rademacher complexity \cite{khim2019adversarial, DBLP:journals/corr/abs-1810-11914, DBLP:journals/corr/abs-2004-13617, xiao2022adversarial, DBLP:journals/corr/abs-1810-02180}, VC dimension \cite{DBLP:journals/corr/abs-1902-04217} and algorithmic stability analysis \cite{xing2021on, xiao2022stability}, and by investigating the problem under the PAC learning frameworks \cite{cullina2018paclearning, DBLP:journals/corr/abs-1906-05815}. 

The robust generalization has also been theoretically explored beyond the perspective of sample complexity. The work of \cite{li2022robust} explains robust generalization from the viewpoint of neural network expressive power. They show that the expressive power of practical models may be inadequate for achieving low robust test error. The work in \cite{DBLP:journals/corr/abs-1906-02931} attempts to understand robust generalization by exploring the inductive bias in gradient descend under the adversarial training setup. Another line of work attempt to understand the overfitting in AT by linking AT with distributional robust optimization (DRO) \cite{kuhn2019wasserstein, sinha2020certifying} . The work of \cite{staib2017distributionally} and \cite{bui2022unified} demonstrate that different AT schemes can be reformulated as special cases in DRO. The work of \cite{bennouna2023certified} shows that under a saddle-point assumption, AT will always cause an at least larger overfitting gap than directly solving an ERM using the standard loss on the data that are adversarially perturbed w.r.t the model obtained by AT.
Numerous endeavors have been undertaken to address the challenge of robust overfitting with various empirical training algorithms proposed. The paper \cite{bai2021recent} and \cite{qian2022survey} provide a comprehensive overview of the latest developments in empirical research in this field.

\section{Adversarial training and induced distributions}

We consider a classification setting with input space ${\mathcal X}\subseteq \mathbb{R}^{d}$ and  label space ${\mathcal Y}:=\{1,2,\cdots, K\}$. 
Let $\Theta$ be the parameter space of a model architecture of interest, and for each  $\theta\in \Theta$, let $l_{\theta}:\mathcal{X}\times\mathcal{Y}\to \mathbb{R}_{+}$ denote a loss function (e.g, the cross-entropy loss) associated with the model $f_\theta$ with parameter $\theta$, where $\ell_\theta(v, y)=0$ if and only if $f_\theta(v)=y$.

For any data distribution $\mathcal {D}$ on
$\mathcal{X}\times\mathcal{Y}$, the standard population error 
$R_{\cal D}(\theta)$ is defined as 
\begin{equation}
\label{def:standard pop risk}
R_{\mathcal{D}}(\theta):=\mathbb{E}_{(x, y)\sim \mathcal{D}}\left[l_{\theta}(x,y)\right]
\end{equation}
and the adversarial population error $R_{\mathcal{D}}^{\rm adv}(\theta)$ is then
\begin{equation}
\label{def:rob pop risk}
R_{\mathcal{D}}^{\rm adv}(\theta):=\mathbb{E}_{(x, y)\sim \mathcal{D}}\left[\max\limits_{v\in\mathbb{B}(x,\epsilon)}l_{\theta}(v,y)\right]
\end{equation}
where $\mathbb{B}(x,\epsilon)$ is customarily chosen as the $\infty$-norm ball centered at $x$ with radius $\epsilon$, or $\mathbb{B}(x,\epsilon):=\{t\in\mathbb{R}^{d}: \|t-x\|_{\infty}\le \epsilon\}$. For later use, we will denote by ${\mathbb U}(x, \epsilon)$ the uniform distribution over ${\mathbb B}(x, \epsilon)$.

In adversarial training, the ultimate objective is to find a model parameter $\theta$ that minimizes $R_{\mathcal{D}}^{\rm adv}(\theta)$. Having no access to $\mathcal {D}$, in practice, a natural choice is to minimize the empirical version of $R_{\mathcal{D}}^{\rm adv}(\theta)$, namely,  
\begin{equation}
\label{eq:rob emp risk} 
R_{S}^{\rm adv}(\theta):=\frac{1}{m}\sum\limits_{i=1}^{m}\max\limits_{v_{i}\in\mathbb{B}(x_i,\epsilon)}l_{\theta}(v_i,y_i)
\end{equation}
on a training set $S:=\{(x_i,y_i)\}_{i=1}^{m}$ drawn i.i.d from $\mathcal{D}$. The most popular approach to solving this problem for neural networks is iterating between solving the inner maximization via $k$-step projected gradient descend (PGD) and updating $\theta$ through stochastic gradient descent. We now give a concise explanation of this procedure, referred to as PGD-AT, restating the procedure in \cite{madry2019deep}. 

\noindent {\bf $k$-step PGD} 
A $k$-step PGD can be described by $k$-fold composition of an one-step PGD mapping. With a fixed choice of $x\in{\mathcal X}$, $y\in{\mathcal Y}$, $\theta\in \Theta$, the one-step PGD mapping ${\mathcal A}_{x, y, \theta}: \mathbb{R}^{d}\to \mathbb{B}(x,\epsilon)$ is defined as
\begin{equation}
\label{eq: 1-pgd}
    {\mathcal A}_{x, y, \theta}(x'):= \Pi_{\mathbb{B}(x,\epsilon)}\left[x' + \lambda\mathrm{sgn}\left(\nabla_{x'}l_{\theta}(x',y)\right)\right] 
\end{equation}
Here $\Pi_{\mathbb{B}(x,\epsilon)}:\mathbb{R}^{d}\to \mathbb{B}(x,\epsilon)$ denotes the operation of projecting onto the set $\mathbb{B}(x,\epsilon)$ and $\lambda\in\mathbb{R}_{+}$ is a hyperparameter. The sgn($\cdot$) function returns the sign of its input, and acts on a vector element-wise. The $k$-step PGD mapping ${\mathcal Q}_{x, y, \theta}: \mathbb{R}^{d}\to \mathbb{B}(x,\epsilon)$ is then
\begin{equation}
\label{eq: k-pgd}
 {\mathcal Q}_{x, y, \theta}(x'):=\left(\underbrace{{\mathcal A}_{x, y, \theta} \circ\cdots\circ{\mathcal A}_{x, y, \theta}}_{k\ \rm times}\right)(x')
\end{equation}

We denote $\{{\cal Q}_{x, y, \theta}: (x, y) \in {\cal X}\times {\cal Y}\}$ collectively by ${\cal Q}_\theta$.

\noindent {\bf Iterations of PGD-AT} At the $t^{\rm th}$ iteration of PGD-AT, where the model parameter is $\theta_t$, the
solution of the inner maximization $\max_{v_i\in {\mathbb B}(x_i, \epsilon)}l_{\theta_t}(v_i, y_i)$ is taken as $l_{\theta_t}(
{\cal Q}_{x_i, y_i, \theta_t}(x_i), y_i)$, and the model parameter is updated, with learning rate $\eta$, by
\begin{equation}
\label{eq:PGD_update}
    \theta_{t+1} = \theta_{t}-\eta\nabla_{\theta_t}\left[\frac{1}{m}\sum\limits_{i=1}^{m} l_{\theta_t}\left({\cal Q}_{x_i, y_i, \theta_t}(x_i), y_i\right)\right]
\end{equation}

If $(X, Y)$ is drawn from ${\cal D}$, then we denote by $\tilde{\cal D}_{\theta_t}$,  or simply $\tilde{\cal D}_t$, the distribution of $({\cal Q}_{X, Y, \theta_t}(X), Y)$, and refer to it as the {\em induced distribution} at iteration $t$. It is clear that 
the update equation in (\ref{eq:PGD_update}) is essentially one-step gradient descent of the standard notion of empirical risk on a training set drawn from $\tilde{\cal D}_t$. The dynamics of the gradient descent is however complicated by the fact that $\tilde{\cal D}_t$ continuously evolves with $t$, due to the dependency of $\mathcal{Q}_{\theta_t}$ on $\theta_t$. The thrust of this work is to investigate if the evolution trajectory of  $\tilde{\cal D}_t$  along PGD-AT iterations may be related to robust overfitting. 

\section{Training on the induced distributions}

The following experiments are conducted. First PGD-AT is performed on a training set $S$. Along this process, for a prescribed set of training iterations (or ``checkpoints'') $I:=\{k_1, k_2, \ldots, k_N\}$, the model parameter $\theta_{t}$ at each checkpoint $t\in I$ is saved. Then at each checkpoint $t$, each example in the training set $S$ is perturbed according to ${\cal Q}_{\theta_t}$, giving rise to the perturbed training set $\tilde{S}_{t}$. The testing set $T$ is also similarly perturbed, giving rise to the perturbed testing set $\tilde{T}_{t}$. The model is then retrained fully on $\tilde{S}_{t}$, using standard training (i.e, without perturbation) with random initialization, where we denote the resulting learned model parameter by $\phi_t$. The model $\phi_t$ is then tested on $\tilde{T}_{t}$. Note that in this setting, both $\tilde{S}_{t}$ and $\tilde{T}_{t}$ are i.i.d. samples from $\tilde{\cal D}_{t}$. We call these experiments ``induced distribution experiment" (IDE) for the ease of reference.
\begin{figure}[!htbp]
    \centering
    \subfigure[CIFAR-10 with PRN-18]{\includegraphics[width=0.24\textwidth]{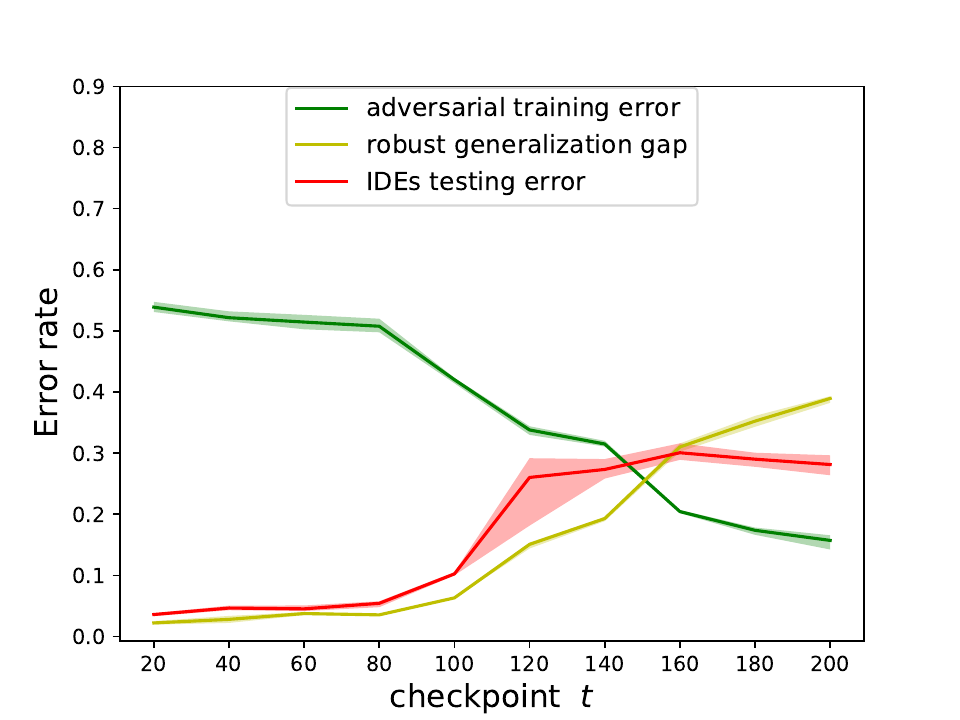}}
    \subfigure[CIFAR-100]{\includegraphics[width=0.24\textwidth]{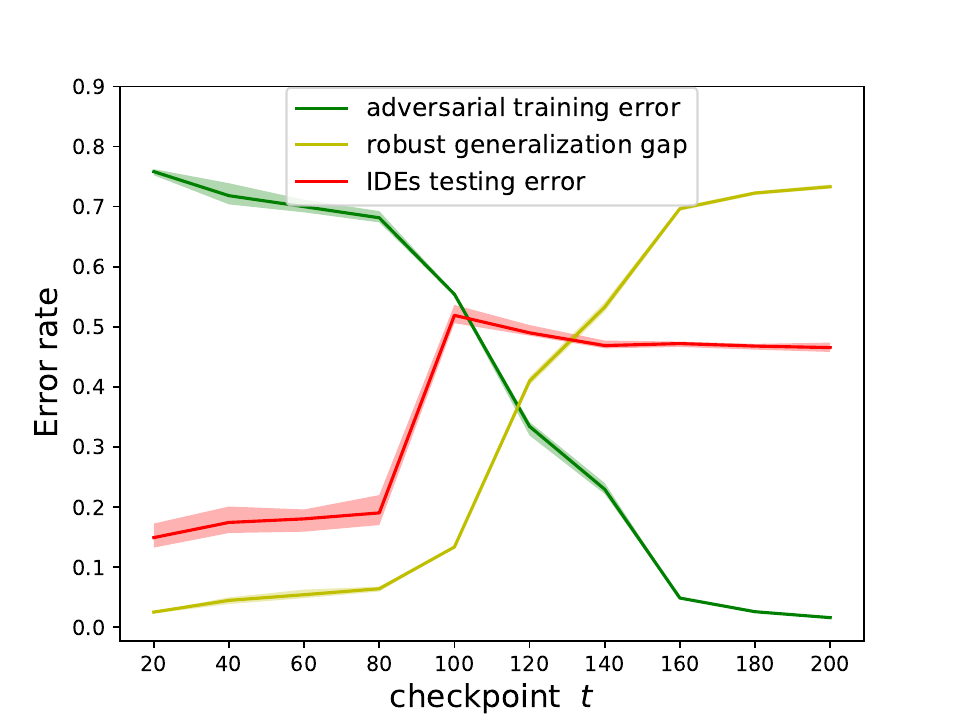}}
    \subfigure[Reduced Imagenet]{\includegraphics[width=0.24\textwidth]{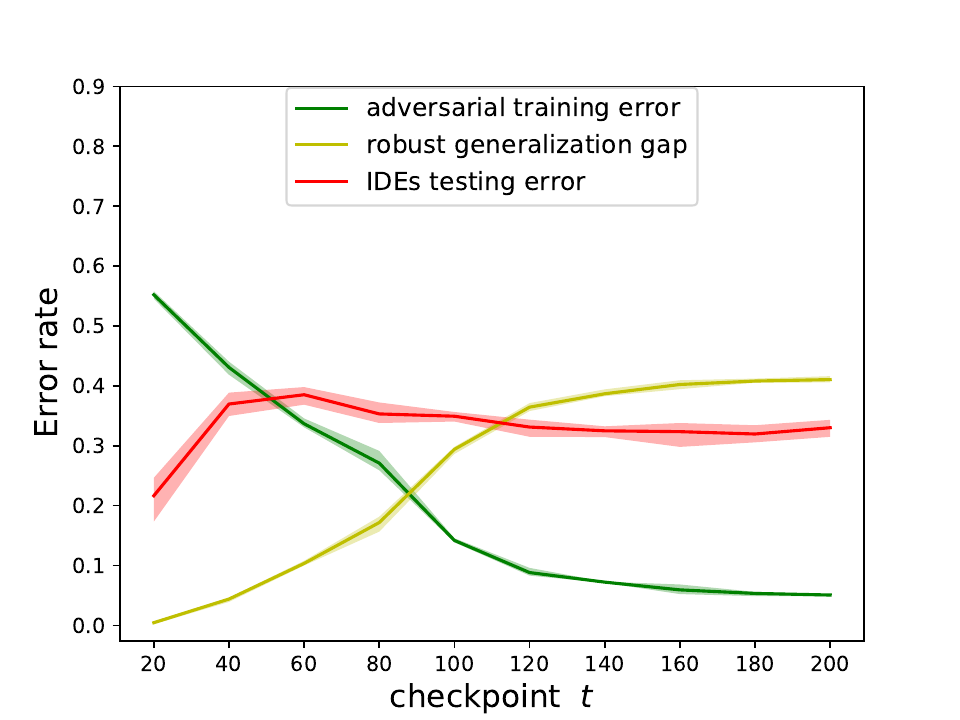}}
    \subfigure[MNIST]{\includegraphics[width=0.24\textwidth]{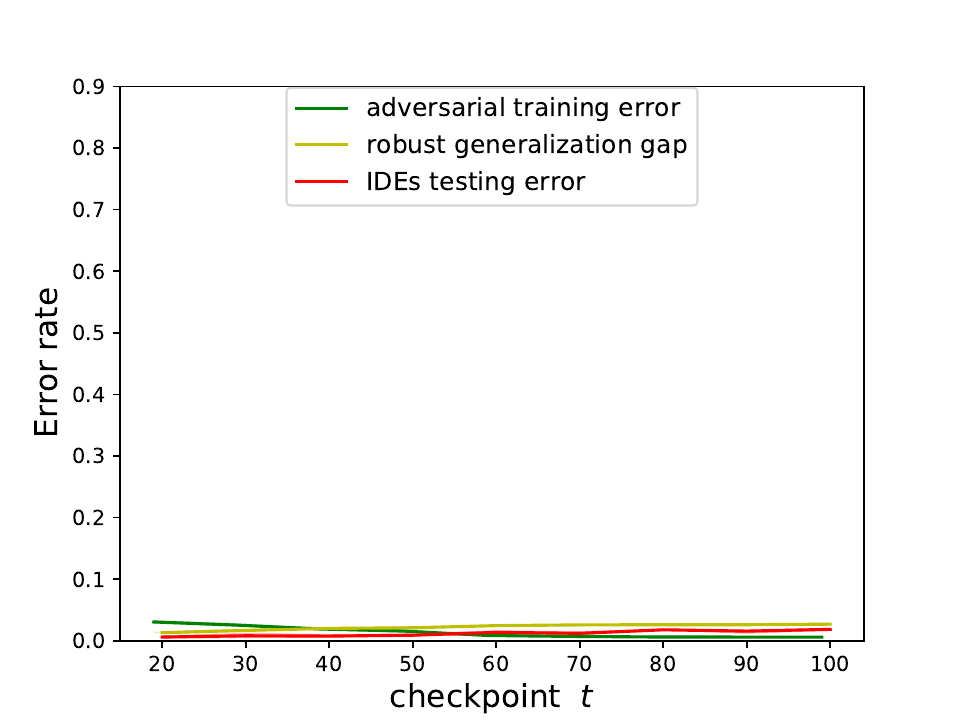}}
    \caption{PGD-AT and the corresponding IDE results across different datasets. The behaviour of the red curves matches that of the yellow curve, as we observe that in (a)-(c) a substantial rise in IDE testing errors concurrent with the emergence of robust overfitting and that in the sub-figure (d) the absence of robust overfitting also coincides with the consistently low IDE testing error. These demonstrate a compelling correlation between these two quantities.}
     \label{fig:IDE cifar100 & resimg}
\end{figure}

The experiments are conducted on MNIST \cite{lecun1998gradient}, CIFAR10 and CIFAR100 \cite{krizhevsky2009learning}. We also conduct experiments on a "scaled-down" version of the ImageNet dataset \cite{russakovsky2015imagenet}. Since PGD-AT is known to be significantly challenging and computationally expensive on the full-scale ImageNet dataset, we draw inspiration from the approach presented in \cite{tsipras2019robustness} and made a Reduced ImageNet by aggregating several subsets of the original ImageNet. Our reduced ImageNet comprises 10 classes, each containing 5000 training samples and approximately 1000 testing samples per class. More details concerning this dataset are given in Appendix \ref{section: setup}. We use the following settings for PGD-AT: For MNIST, following the settings in \cite{madry2019deep}, we train a small CNN model using 40-step PGD with step size $\lambda=0.01$ and perturbation radius $\epsilon=0.3$. For the other three datasets, we train the pre-activation ResNet (PRN) model \cite{DBLP:journals/corr/HeZR016} and the Wide ResNet (WRN) model \cite{DBLP:journals/corr/ZagoruykoK16}. We use 
5-step PGD with $\epsilon=4/255$ for the Reduced ImageNet and 10-step PGD with $\epsilon=8/255$ for CIFAR-10 and CIFAR-100 according to \cite{DBLP:journals/corr/abs-2002-11569} in PGD-AT. We set $\lambda=2/255$ on CIFAR10 and CIFAR100,  $\lambda=0.9/255$ on the reduced ImageNet. More details concerning the hyperparameter settings are given in Appendix \ref{section: setup}.
\begin{figure}[!htpb]
    \centering
    \includegraphics[width=0.32\textwidth]{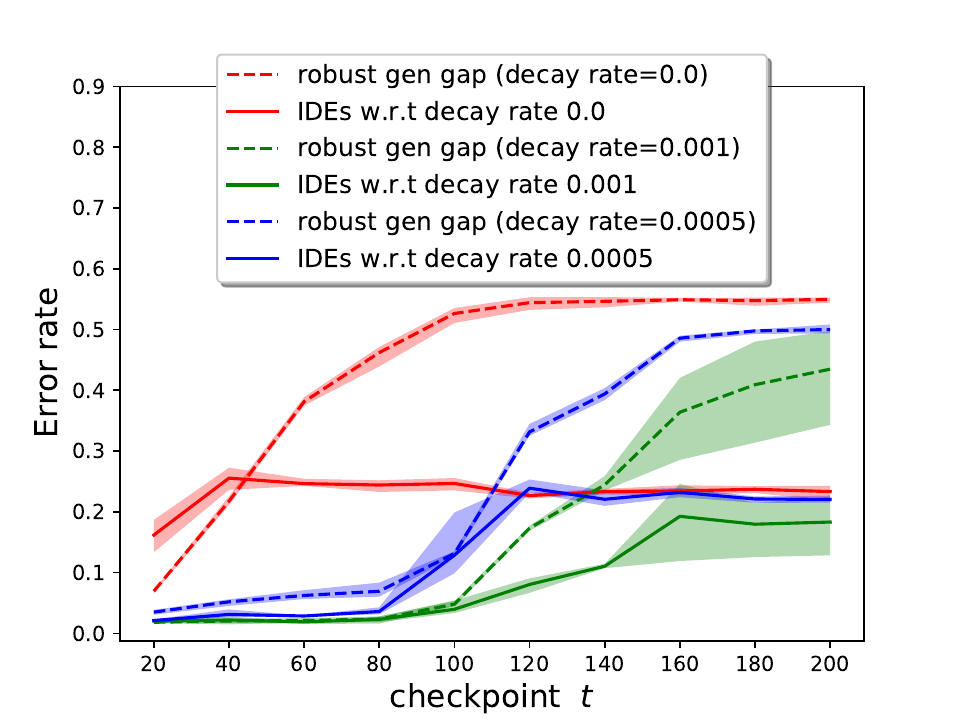}
    \caption{The outcomes of additional experiments conducted on CIFAR-10. In the experiments, we perform PGD-AT with various weight decay rates and conduct IDEs for each of the PGD-AT variant. The blue curves are reproduced from Figure \ref{fig:IDE cifar10}, serving as a reference for a clear comparison. The results further solidify the correlation between the robust overfitting and the IDE testing error.}
    \label{fig:IDE mnist & wd}
\end{figure}

For each dataset, the experiments are repeated five times with different random seeds. The experimental results on CIFAR-10 and CIFAR-100 are shown in Figure \ref{fig:IDE cifar10}, \ref{fig:IDE cifar100 & resimg} (a) and (b), where the green and yellow curves illustrate a phenomenon known as robust overfitting  \cite{DBLP:journals/corr/abs-2002-11569}: after a certain point in PGD-AT, the robust generalization gap (yellow curves), that is $|R_{\mathcal{D}}^{\rm adv}(\theta_t)-R_{S}^{\rm adv}(\theta_t)|$, steadily increases while the adversarial training error (green curves) constantly decreases. The red curves in the figures depict the testing errors of IDEs w.r.t various checkpoints in PGD-AT. Notably, a significant rise in the average IDE testing error is observed between the 80th and 120th checkpoints, increasing from 3.6\% to 23.89\% for CIFAR-10 and from 19\% to 48.99\% for CIFAR-100. Furthermore, this shift coincides with the onset of robust overfitting, where a significant rise in robust generalization gap is also observed. These results indicate that as $\theta_t$ evolves along PGD-AT, the model trained on the samples of $\tilde{\mathcal{D}}_t$ becomes harder to generalize. More importantly, this difficulty in generalization appears to be closely linked to the robust overfitting phenomenon. This potential connection is further demonstrated by the experimental results on the reduced ImageNet (see Figure \ref{fig:IDE cifar100 & resimg} (c)), where robust overfitting emerges at an earlier training stage and simultaneously a rise in the IDE testing error occurs. This increment in the IDE testing error is also substantial, with an averaged error of  21.65\% at the 20th checkpoint elevating to 38.52\% at the 60th checkpoint.

Our experiments on MNIST (see Figure \ref{fig:IDE cifar100 & resimg} (d)) exhibits a scenario where a good robust generalization is achieved.  Interestingly, a small IDE testing error is maintained throughout the evolution of $\tilde{\mathcal{D}}_t$ with the absence of robust overfitting. Figure \ref{fig:IDE mnist & wd}  shows results from additional experiments on CIFAR-10. In these experiments, we perform PGD-AT with different level of weight decay to control the level of robust overfitting. Subsequently, IDEs are conducted for each such variant of PGD-AT. In Figure \ref{fig:IDE mnist & wd}, each distinct color corresponds to a different weight decay factor utilized in PGD-AT. Within each color category, the dashed curves and the corresponding solid lines represent, respectively, the robust generalization gaps and the IDE results associated with that specific PGD-AT variant. From these results, we see that increasing the weight decay factor results in a notable reduction in the robust generalization gap, while conversely, decreasing the weight decay factor leads to the opposite effect. This is shown by the downward shift in the dashed curves across the three color categories. More noteworthy is a clear synchronization observed between each pair of dashed and solid curves, with lower dashed curves consistently corresponding to lower solid curves in the same color category.

All these results suggest a strong correlation between the IDE test error $R_{\tilde{\cal D}_t}(\phi_t)$ and the robust generalization gap $R_{\cal D}^{\rm adv} (\theta_t) - R_{S}^{\rm adv} (\theta_t)$. Although by construction, $R^{\rm adv}_{\mathcal{D}}(\theta_t)= R_{\tilde{\cal D}_{t}}(\theta_t)$ and $R^{\rm adv}_{S}(\theta_t)= R_{\tilde{S}_{t}}(\theta_t)$, and hence the robust generalization gap can be re-expressed by $R_{\tilde{\cal D}_t} (\theta_t) - R_{\tilde{S}_t} (\theta_t)$, such a correlation is still quite surprising. This is because the learning of the parameter $\phi_t$ has been started from a completely random initialization and one would not expect the resulting parameter $\phi_t$ is linked to the parameter $\theta_t$ in any obvious way, despite that the latter contributes to shaping  the distribution $\tilde{\cal D}_t$. 

A novel observation in this work, this correlation, is certainly curious in its own right and deserves further investigation. At this point, it has at least highlighted the impact of the dynamics of PGD-AT on robust overfitting, beyond the static quantities, such as loss landscape. It also suggests that the increasing  difficulty of generalization inherent in the induced distribution plays an important role in robust overfitting. 

\section{Generalization properties of the induced distributions}
The experiments above reveals an interesting phenomenon that as $\tilde{\mathcal{D}}_t$ evolves in PGD-AT, the model obtained from $\tilde{\mathcal{D}}_t$ becomes harder to generalize, particularly when robust overfitting occurs. This suggests that the increasing generalization difficulty of the induced distribution $\tilde{\cal D}_t$ along PGD-AT trajectory contributes as an important factor to robust overfitting. It remains curious what causes $\tilde{\cal D}_t$ to become harder to generalize in adversarial training. Here we provide a theoretical explanation and corroborate it with empirical observations.

There are two views of adversarial examples, ``in-distribution view'' (e.g., in \cite{DBLP:journals/corr/abs-1710-10766, gilmer2018adversarial, raghunathan2019adversarial}) and "out-of-distribution view'' (e.g., in \cite{szegedy2014intriguing, khoury2018geometry,  stutz2019disentangling}). In the out-of-distribution view,  adversarial examples are considered as living outside the data manifold. In the in-distribution view, adversarial examples are considered as located within the support of the true data distribution but having low probability (density). This paper takes the in-distribution view and, instead of regarding the data distribution ${\cal D}$ as ``absolutely robust'', we consider it only has a weaker notion of robustness guarantee.  We now make precise this notion and our assumption of ${\cal D}$. 

Underlying the data distribution ${\cal D}$, we assume that there exists another distribution ${\cal D}^*$ on ${\cal X}\times {\cal Y}$. We may re-express ${\cal D}^*$ as the pair $({\cal D}^*_{\cal X}, h^*)$, where ${\cal D}^*_{\cal X}$ is the marginal distribution of ${\cal D}^*$ on the input space ${\cal X}$ and $h^*$ is the ground-truth classifier, a function mapping ${\cal X}$ to ${\cal Y}$. The distribution ${\cal D}^*$ satisfies the following property: when $X$ is drawn from ${\cal D}^*_{\cal X}$ and $\rho$ is drawn independently of $X$ from ${\mathbb U}(0, \epsilon)$, $h^*(X+\rho)=h^*(X)$ with probability 1. That is, the classifier $h^*$ is absolutely robust (against adversarial perturbations with norm no larger than $\epsilon$) when $X$ is drawn from ${\cal D}^*_{\cal X}$ -\textemdash in this case, we also say, in short, ``${\cal D}^*$ is absolutely robust''. We however note that in complex learning tasks,  distributions with such strong notion of robustness may not arise naturally as the true data distribution. Instead, we assume that the true data distribution ${\cal D}$ is generated from ${\cal D}^*$ as follows: Draw $X$ from ${\cal D}^*_{\cal X}$, draw $\rho$ independent of $X$ from ${\mathbb U}(0, \epsilon)$, the distribution of $(X+\rho, h^*(X))$ is the true data distribution ${\cal D}$. That is, an instance drawn from the true data distribution is only guaranteed, with high probability, to resist all perturbations with magnitude no more than $\epsilon$. This assumption of data distribution ${\cal D}$ is consistent with the in-distribution perspective of adversarial examples and such weaker notion of robustness guarantee is also reminiscent of the notion of $(\epsilon, \delta)$-robustness recently presented in \cite{epsDeltaRobust:2023}. 

We now study the (standard) classification problem arising in the IDE setting at check point $t$, namely, that with $\tilde{\mathcal{D}}_t$ as the data distribution. Specifically, we wish to understand how difficult is such a learning task. 

Let ${\cal H}$ be a hypothesis class for this learning problem, where each member $h\in {\cal H}$ is a function mapping ${\cal X}$ to ${\cal Y}$. Note that the hypothesis class ${\cal H}$ may have not be related to the model used for adversarial training in any way.  Let $\ell: {\cal Y} \times {\cal Y} \rightarrow {\mathbb R}$ be a non-negative loss function. For every $h\in {\cal H}$, let function $f_h:{\cal X}\times {\cal Y} \rightarrow {\mathbb R}$ be defined by $f_h(x, y):= \ell(h(x), y)$. Let ${\cal F}:=\{f_h: h\in {\cal H}\}$. From here on, we will restrict our attention to this ``loss hypothesis class'' ${\cal F}$ and study how well the members of ${\cal F}$ generalize from an i.i.d. training sample 
$\{(v_i, y_i)\}_{i=1}^{m}$ drawn  from $\tilde{\mathcal{D}}_t$ to the unknown distribution $\tilde{\mathcal{D}}_{t}$.  Specifically, for each $f\in {\cal F}$, the key quantity of our interest is the {\em generalization gap} of each  $f$ with respect to $\tilde{\cal D}_t$, defined by
 \[
 {\rm GG}_t(f):=
 \left|\frac{1}{m}\sum\limits_{i=1}^{m}f(v_i,y_i)-\mathbb{E}_{(v,y)\sim \tilde{\cal D}_t} f(v,y)\right|.
 \]

As we will soon show, the generalization gap turns out to be related to a key quantity characterizing a local property of the perturbation map $Q_{x, y, \theta_t}$, through which $\tilde{\cal D}_t$ is induced.

Let $\rho$ and $\rho'$ be drawn independently from ${\mathbb U}(0, \epsilon)$, and define

\begin{equation}
\label{eq:localDispersion}
\tilde{\gamma}_{t}(x, y):=\mathbb{E}_{\rho, \rho'} \|\mathcal{Q}_{x,y,\theta_t}(x+\rho)-\mathcal{Q}_{x,y,\theta_t}(x+\rho')\|_{2}^{2}.
\end{equation}

We refer to this quantity as the {\em local dispersion} of ${\cal Q}_{x, y, \theta_t}$, as it measures the spread of the perturbation of a random point in ${\mathbb B}(x, \epsilon)$ by ${\cal Q}_{x, y, \theta_t}$ \textemdash To see this, one may verify that $\tilde{\gamma}_{t}(x, y)$ can be expressed as 
\[
\tilde{\gamma}_t(x, y)= 2 \cdot {\rm Trace}\left({\rm COV}({\cal Q}_{x, y, \theta_t}(V))\right)
\]
where $V$ is drawn from ${\mathbb U}(x, \epsilon)$ and ${\rm COV}({\cal Q}_{x, y, \theta_t}(V))$ denotes the covariance matrix of the random vector ${\cal Q}_{x, y, \theta_t}(V)$. 

One may argue intuitively that smaller local dispersion of ${\cal Q}_{\theta_t}$ may allow the resulting distribution $\tilde{\cal D}_t$ to generalize better: consider an instance $(X, Y)$ drawn from the absolutely robust distribution ${\cal D}^*$, and two data points
$(X+\rho, Y)$ and $(X+\rho', Y)$ (with $\rho$ and $\rho'$ drawn independently from ${\mathbb U}(0, \epsilon)$.  Suppose that $(X+\rho, Y)$ is included in the training set and $(X+\rho', Y)$
is included in the testing set. When the local dispersion $\tilde{\gamma}_t(X, Y)$ is small, the perturbed version of  training point $(X+\rho, Y)$ and that of the testing point $(X+\rho', Y)$ (both of which are realizations from $\tilde{\cal D}_t$) are close, allowing the latter to behave similarly as the former. 

We now rigorously formalize this intuition, under the following assumptions. 
\begin{itemize}
\item (Lipchitzness of $f$ over ${\cal X}$)
For any $y\in\mathcal{Y}$,  $|f(x,y)-f(x',y)|\le \beta\|x-x'\|_2$ for $\forall x,x'\in \mathcal{X}$. 
\item (Loss boundedness) $\sup\limits_{x,y\in\mathcal{X}\times{\mathcal{Y}}}|f(x,y)|= B<\infty$. 

\end{itemize}

\begin{theorem} Let $f \in {\cal F}$ and 
suppose that $f$ satisfies the above conditions. Then for any $\tau>0$, with probability at least $1-\tau$ over the i.i.d. draws of sample $\{(v_i, y_i)\}_{i=1}^{m}$ from $\tilde{\mathcal{D}}_{t}$, 

\begin{equation}
\label{eq: bound}
    {\rm GG}_t(f)
    \le \frac{2\beta}{\sqrt{m}}\sqrt{\mathbb{E}_{{\mathcal{D}^{*}}}\tilde{\gamma}_{t}(x,y)} + \frac{2\beta\sqrt{d}\epsilon}{\sqrt{m}}+ \frac{2B}{\sqrt{m}}\left(\sqrt{\frac{\log\frac{1}{\tau}}{2}}+1\right)
\end{equation}
\label{main theorem}
\end{theorem}

The derivation of the bound is based on a modification of the Rademacher complexity analysis. To directly analyze the Rademacher complexity of the hypothesis class ${\cal F}$ is however difficult in our setting due to the lack of restrictions on ${\cal F}$. A future direction towards our work is extending our analysis to more restricted hypothesis class. We leave the proof of the Theorem in Appendix \ref{section: proof}. The theorem shows that the generalization gap of any $f$ w.r.t to the distribution $\tilde{\mathcal{D}}_{t}$ is affected by the 
expected local dispersion (ELD)
$\mathbb{E}_{{\mathcal{D}^{*}}}\tilde{\gamma}_{t}(x,y)$ 
of ${\cal Q}_{\theta_t}$. The key message of this theorem is that a small generalization gap is guaranteed when  $\mathbb{E}_{{\mathcal{D}^{*}}}\tilde{\gamma}_{t}(x,y)$ is small. 
\begin{figure}[!htpb]
    \centering
    
    \subfigure[CIFAR-10]{\includegraphics[width=0.23\textwidth]{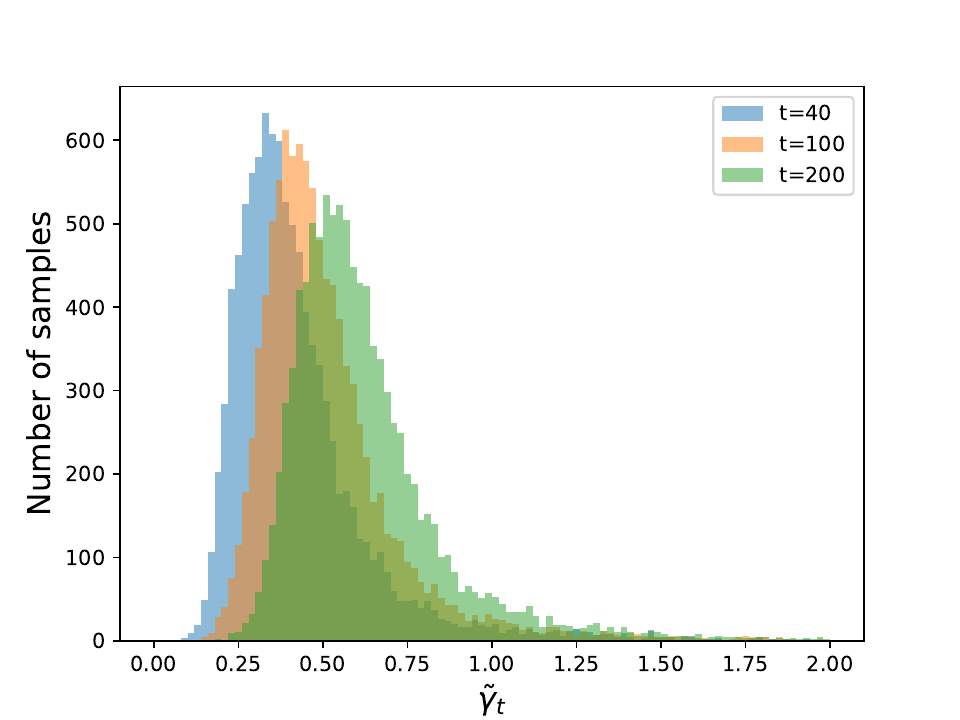}}
     \subfigure[CIFAR-100]{\includegraphics[width=0.23\textwidth]{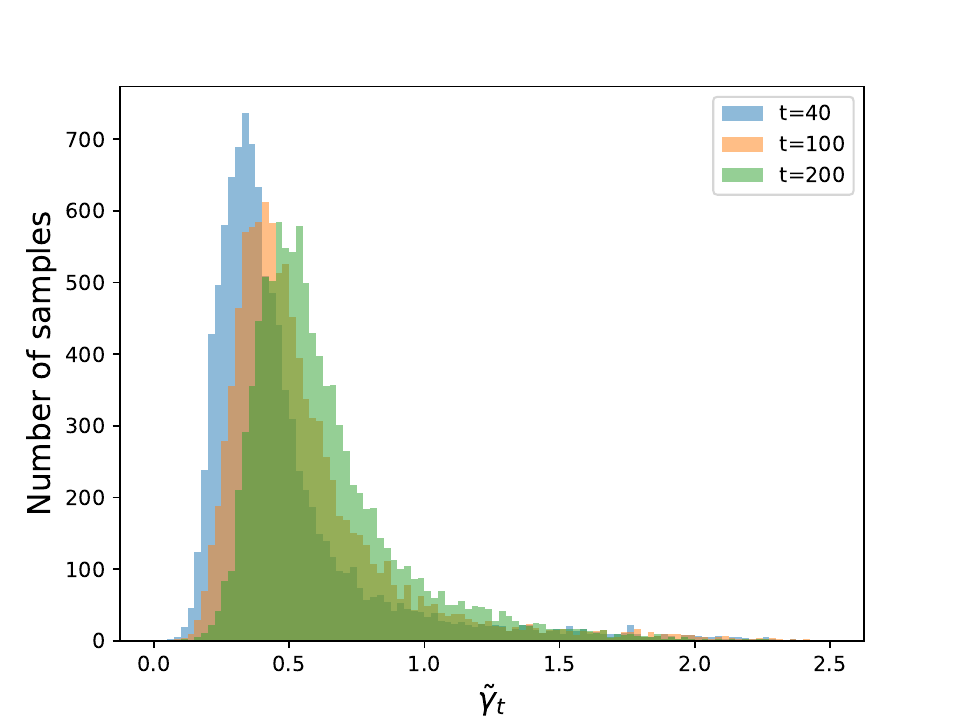}}
    \subfigure[CIFAR-10]{\includegraphics[width=0.23\textwidth]{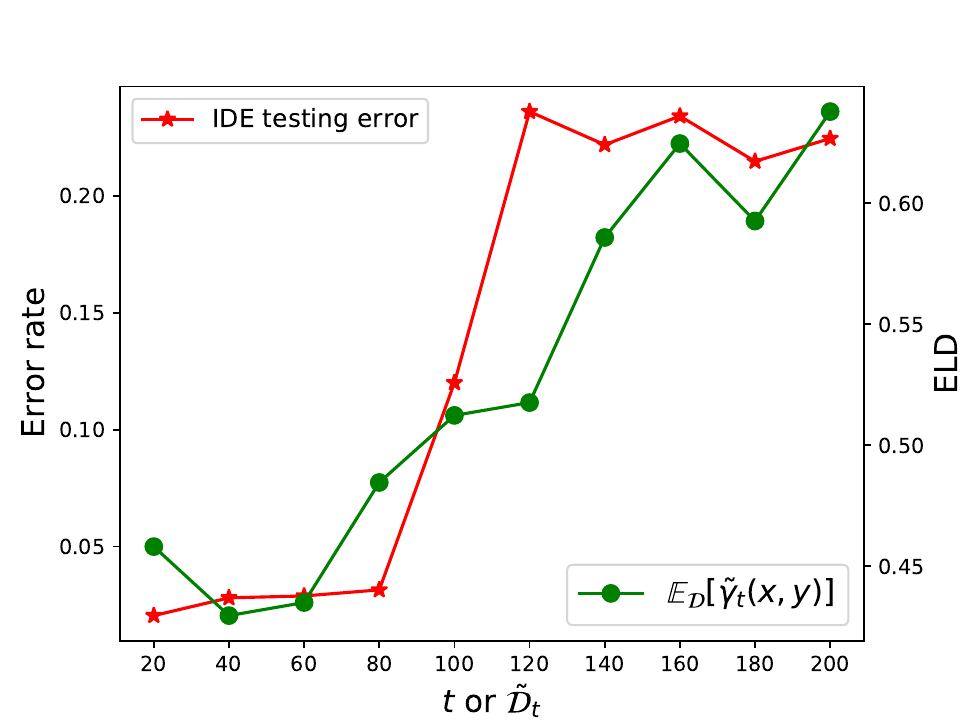}}
     \subfigure[CIFAR-100]{\includegraphics[width=0.23\textwidth]{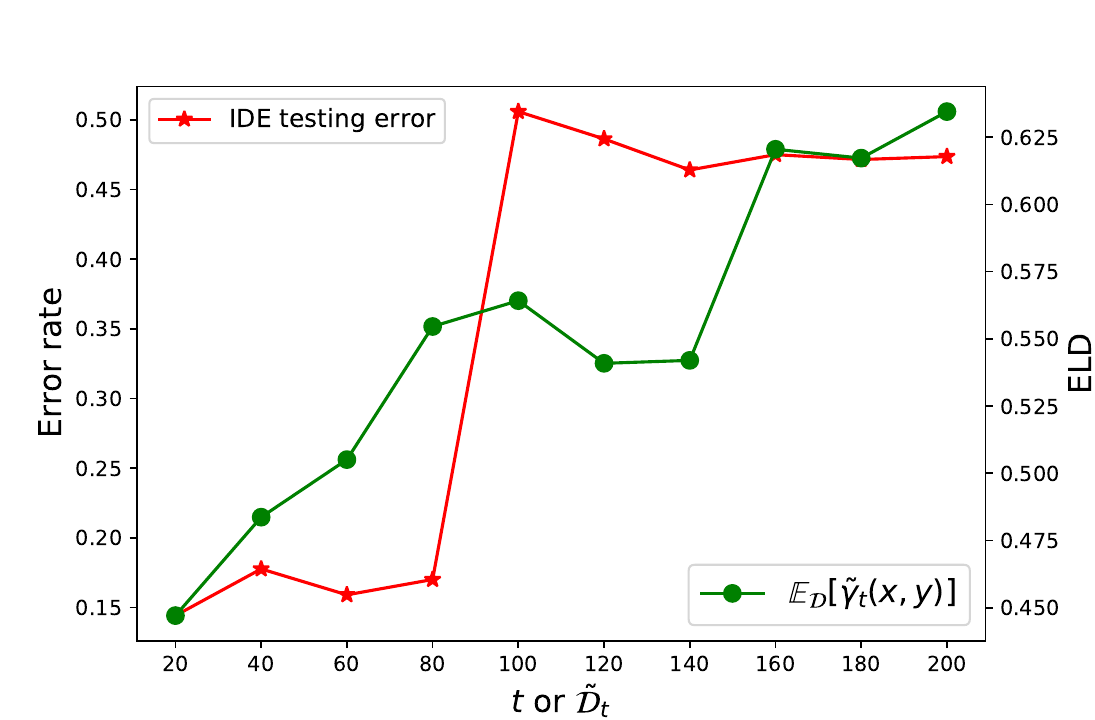}}
 
     \caption{Local dispersion measured on the CIFAR-10 and CIFAR-100 testing set. (a) and (b): histograms of $\tilde{\gamma}_{t}(x,y)$ at three distinct PGD-AT checkpoints. (c) and (d): The evolution of ELD  w.r.t $t$ and the IDE testing error for each $\tilde{\mathcal{D}}_t$. The results show that the level of $\tilde{\gamma}_{t}(x,y)$ increases during PGD-AT and correspondingly the model obtained from $\tilde{\mathcal{D}}_t$ becomes harder to generalize. This implies that the local property of $\mathcal{Q}_{x,y,\theta_t}$ characterized by $\tilde{\gamma}_{t}(x,y)$ plays a dominate role in influencing the generalization of $\tilde{\mathcal{D}}_t$.}
\label{fig: disper}
\end{figure}
 
To validate this theorem, we performed experiments to estimate the ELD for $\theta_t$ at various checkpoints $t$. Note that the expectation here is over the distribution ${\cal D}^*$, from which no samples are available. However, due to the relationship between ${\cal D}_{\cal X}$ and ${\cal D}^*_{\cal X}$, namely that ${\cal D}_{\cal X}$ is  merely an $\epsilon$-smoothed version of ${\cal D}^*_{\cal X}$, one expects that when we draw $x$ from ${\cal D}$, ${\cal D}_{\cal X}(x) \approx {\cal D}^*_{\cal X}(x)$ with high probability. Then when we estimate ${\mathbb E}_{{\cal D}^*} \tilde{\gamma}_t(x, y)$ using an i.i.d. sample $S$ drawn from ${\cal D}$, we may approximately regard $S$ as drawn i.i.d. from  ${\cal D}^*$, and estimate ${\mathbb E}_{{\cal D}^*} \tilde{\gamma}_t(x, y)$ by
 $
 {\mathbb E}_{{\cal D}^*} \tilde{\gamma}_t(x, y) \approx \frac{1}{m} \sum_{i=1}^m  \tilde{\gamma}_t(x_i, y_i),
 $
 where $\{(x_i, y_i)\}_{i=1}^{m}$ are drawn from ${\cal D}$. Notably this is a slightly biased estimator of ${\mathbb E}_{{\cal D}^*} \tilde{\gamma}_t(x, y)$, and the bias decreases with $\epsilon$. In our experiments, the estimation of each $\tilde{\gamma}_t(x_i, y_i)$ is done by Monte-Carlo approximation via sampling 250 pairs of $(\rho, \rho')$ and approximating the  expectation in 
(\ref{eq:localDispersion}) by the sample mean. 

We inspect the evolution of the distribution of  $\tilde{\gamma}_{t}(x,y)$ along the PGD-AT trajectory. Figure \ref{fig: disper} (a) and (b) respectively plot the histograms of $\tilde{\gamma}_{t}(x,y)$ for the testing set of CIFAR-10 and CIFAR-100 at three different PGD-AT checkpoints. It is clear that 
the distribution shifts to the right as PGD-AT proceeds, indicating that 
the perturbation operator ${\cal Q}_{\theta_t}$ becomes more locally dispersive as training goes on. 

We then estimate ELD on the testing sets (the green curves in Figure \ref{fig: disper} (c) and (d)). We plot the corresponding IDE results using red curve in each sub-figure as a clearer comparison. As shown in the figures, ELD
is getting larger along PGD-AT, 
correlating with the increasing difficulty of
generalization on $\tilde{\mathcal{D}}_t$. These results confirm that the bound in Theorem \ref{main theorem} adequately characterizes the generalization behaviour in the IDE experiments, which in turn suggests that the local dispersiveness of the perturbation operator contributes to robust overfitting in PGD-AT. Similar experimental results are also observed across other datasets (see Appendix \ref{section: disper}).

\section{Other observations and discussions}
Our preceding theoretical analysis underscores the critical role played by the local properties of $\mathcal{Q}_{x,y,\theta_t}$ in affecting the generalization performance of models learned on samples from $\tilde{\mathcal{D}}_{t}$. This, in turn, inspires our curiosity to investigate whether additional local properties, beyond local dispersion, also posses critical influences on the generalization of  $\tilde{\mathcal{D}}_{t}$. As such, we inspect the expected distance between the adversarial examples generated by PGD and its clean counterparts, defined as
\begin{equation}
    d_{t}(x,y):= \mathbb{E}_{\rho\sim {\mathbb U}(0, \epsilon) }\|\mathcal{Q}_{x,y,\theta_t}(x+\rho)-x\|_2
\end{equation}

By triangle inequality, it can be easily deduced that when 
$\rho$ and $\rho'$ are drawn independently from ${\mathbb U}(0, \epsilon)$, 
\begin{align}
 & \mathbb{E}_{\rho, \rho'}\|\mathcal{Q}_{x,y,\theta_t}(x+\rho)-\mathcal{Q}_{x,y,\theta_t}(x+\rho')\|_{2} 
   \le  2 d_{t}(x,y) \label{eq: d2c 3}
\end{align}
It is apparent that the term on the left hand side of this inequality is related to the local dispersion of $\mathcal{Q}_{x,y,\theta_t}$ despite that the local dispersion is defined using the square of the $l_2$-distance. This inequality then motivates us to investigate if $d_t(x, y)$ may behave similarly as 
the local dispersion $\tilde{\gamma}_t(x, y)$. 

\begin{figure}[!htbp]
    \centering
    
    \subfigure[CIFAR-10]{\includegraphics[width=0.23\textwidth]{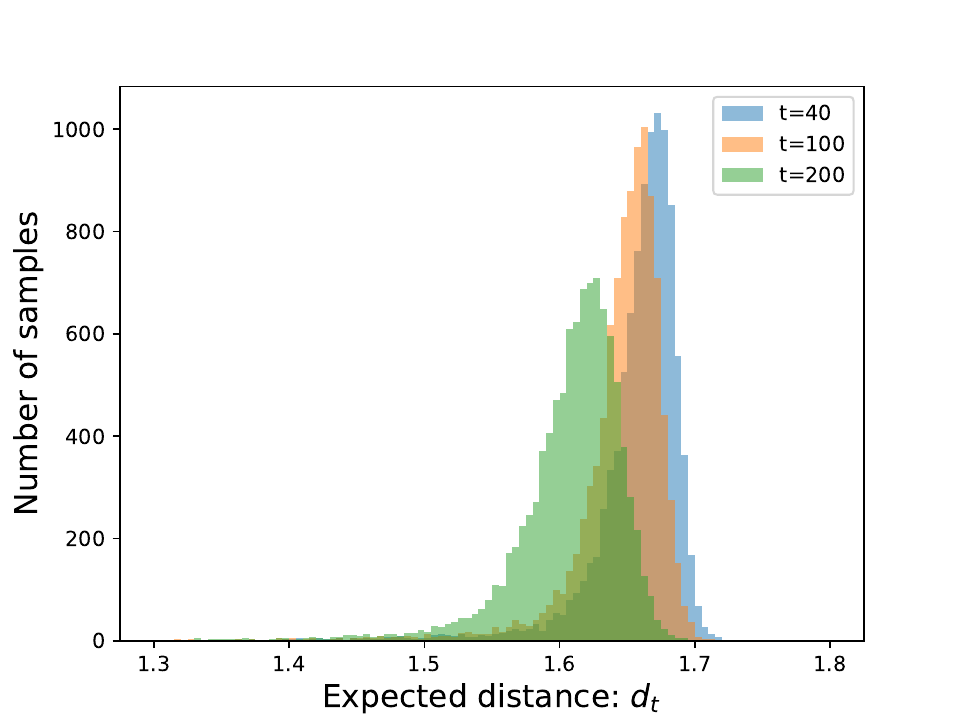}}
     \subfigure[CIFAR-100]{\includegraphics[width=0.23\textwidth]{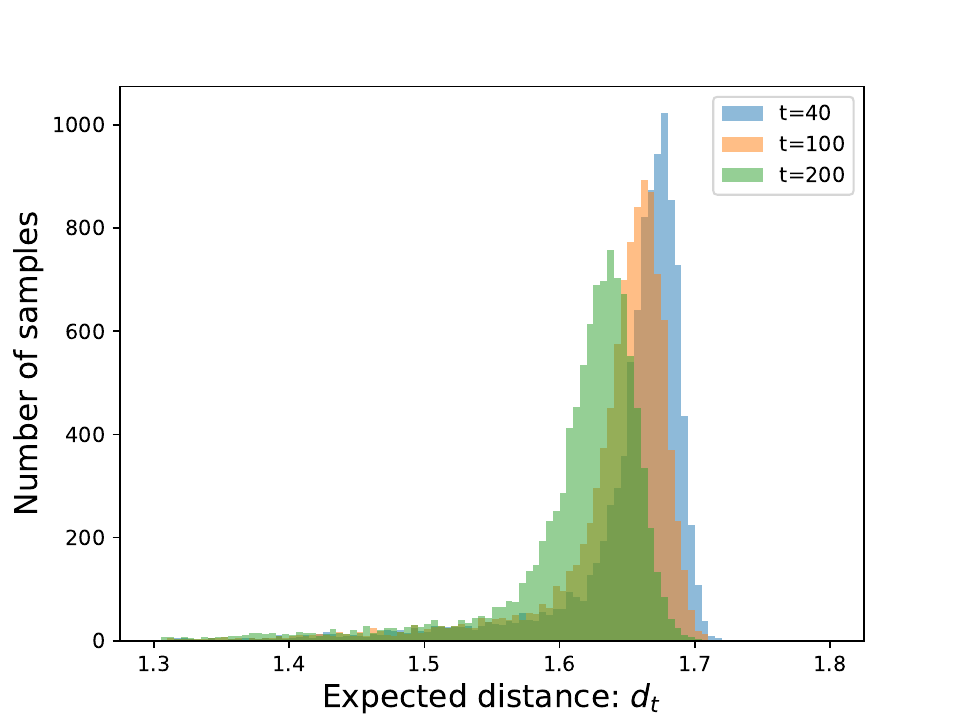}}
    \subfigure[CIFAR-10]{\includegraphics[width=0.23\textwidth]{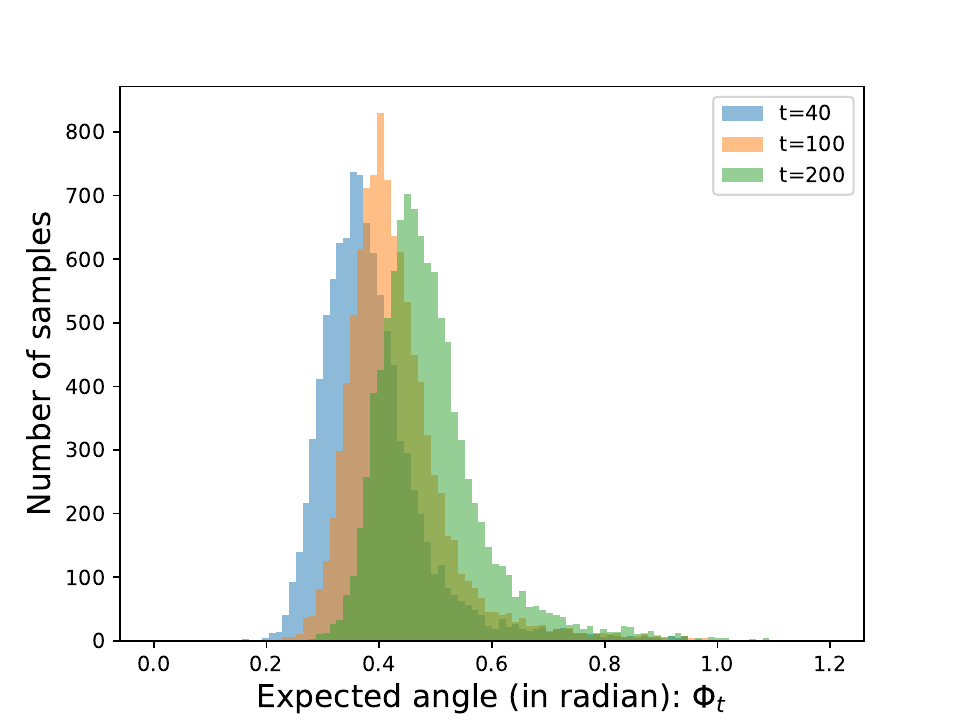}}
     \subfigure[CIFAR-100]{\includegraphics[width=0.23\textwidth]{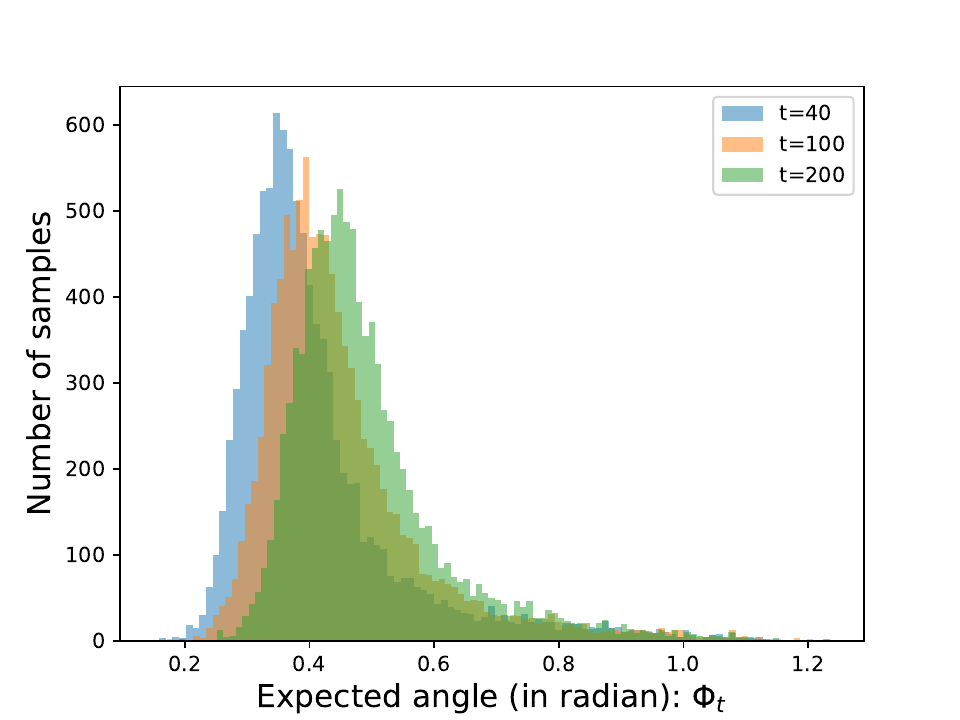}}
    \subfigure[CIFAR-10]{\includegraphics[width=0.23\textwidth]{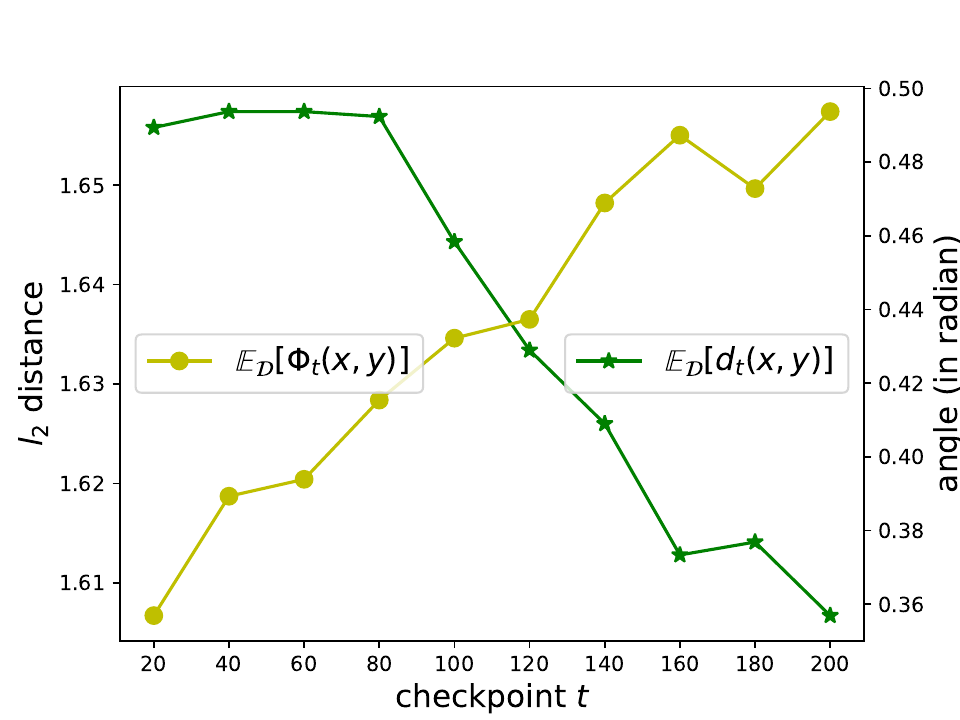}}
    \subfigure[CIFAR-100]{\includegraphics[width=0.23\textwidth]{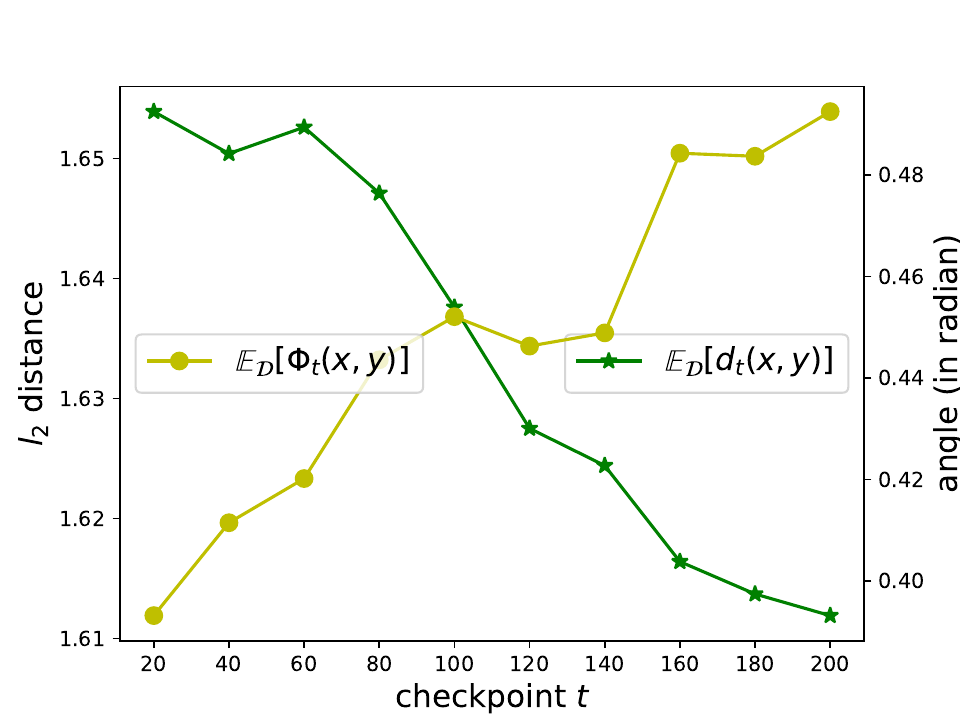}}
 
   \caption{Experiments on the CIFAR-10 and CIFAR-100 testing set. (a) and (b): histograms of $d_{t}(x,y)$. (c) and (d): histograms of $\Phi_t(x,y)$. (e) and (f): The evolution of $\mathbb{E}_{\mathcal{D}}d_{t}(x,y)$ and $\mathbb{E}_{\mathcal{D}}\Phi_t(x,y)$ along PGD-AT trajectory. Combined with the results in Figure \ref{fig: disper}, an interesting phenomenon in PGD-AT is revealed: as the training proceeds, the perturbed data generated by $x$ are getting closer to $x$ and in the meanwhile getting more dispersed potentially due to the spreading of perturbation angles. } 
\label{fig: d2c_angle}
\end{figure}

We conducted Monte-Carlo estimation of $d_{t}(x,y)$ by sampling 250 realizations of $\rho$ drawn from ${\mathbb U}(0, \epsilon)$. We analyze the dynamic behavior of $d_{t}(x,y)$ along the PGD-AT trajectory. In Figure \ref{fig: d2c_angle} (a) and (b), we present histogram of $d_{t}(x,y)$ for the CIFAR-10 and CIFAR-100 testing set at three distinct training checkpoints. Notably, the histogram exhibits a notable mode shift towards a smaller value, indicating a trend that as PGD-AT proceeds, the generated adversarial examples progressively approach their clean counterparts. The reduction in the level of $d_{t}(x,y)$ along PGD-AT is further observed by estimating the expectation $\mathbb{E}_{\mathcal{D}^*}d_{t}(x,y)$ (again, the expectation over
${\cal D}^*$ is approximated by expectation over ${\cal D}$)
on the testing set (see Figure \ref{fig: d2c_angle} (e) and (f), green curves), where a clear drop in $\mathbb{E}_{\mathcal{D}^{*}}d_{t}(x,y)$ is exhibited. From these results, the behavior of $d_t(x,y)$ along PGD-AT shows a drastically different trend. Notably, although these results might seem to suggest that there is strong negative correlation between $\gamma_t(x, y)$ and $d_t(x, y)$, additional observations (see Appendix \ref{section: angle} Figure \ref{fig: disper_d2c}) suggest that this need not be the case. Thus ${\mathbb E}_{{\cal D}^*} d_t(x, y)$ is not as indicative of the IDE generalization gap (and hence of robust overfitting) as ${\mathbb E}_{{\cal D}^*} \tilde{\gamma}_t(x, y)$. 

To develop a deeper understanding of  the difference between $\tilde{\gamma}_t(x, y)$ and $\tilde{d}_t(x, y)$ and the above observations, consider random instance $(x, y)$ drawn from ${\cal D}^*$, which we will refer to as a ``robust anchor'' \textemdash recalling that ${\cal D}^*$ is absolutely robust. Draw $\rho$ from ${\mathbb U}(0, \epsilon)$, and define random variable $\Delta_t(x, y):=
\mathcal{Q}_{x,y,\theta_t}(x+\rho)-x$. It is clear that $d_t(x, y)$ is the mean norm of $\Delta_t(x, y)$ and $\tilde{\gamma}_t(x, y)$ is (up to scale) the trace of the covariance of $\Delta_t(x, y)$, as we shown before. 
Then we see that $d_t(x, y)$ measures, on average, how far the data point $x+\rho$ is away from its robust anchor $x$ after it is perturbed by ${\cal Q}_{x, y,\theta_t}$; thus $d_t(x, y)$ may be seen as the ``bias'' that the perturbation operator introduces to a data point relative to its robust anchor. On the other hand, $\gamma_t(x, y)$ is a notion of ``variance'', measuring, on average, to what extent the data point $x$ is dispersed by the perturbation operator with respect to its robust anchor. The above experimental observations then simply suggest that it is the variance, not the bias, that is closely related to the IDE genearalization gap and to robust overfitting.

As previously noted, when robust overfitting occurs, the local dispersion tends to increase along the PGD-AT trajectory. It is intuitively sensible that increased dispersion may lead to wider spread of the perturbation angles. To formalize this notion of angular spread, we define
\begin{equation}
    \Phi_{t}(x,y):=\mathbb{E}_{\rho, \rho'}\cos^{-1}\left(
    \frac{Z^T_t(x, y) Z'_t(x, y)}{\|Z_t(x,y)\|_2 \|Z_t(x,y)\|_2}   
    \right)
\end{equation}
where $\rho$ and $\rho'$ are drawn independently from ${\mathbb U}(0, \epsilon)$, $Z_t(x,y):= \mathcal{Q}_{x,y,\theta_t}(x+\rho)$, and $Z'_t(x,y):= \mathcal{Q}_{x,y,\theta_t}(x+\rho')$. 

Via a similar Monte-Carlo approach, we may estimate $\Phi_t(x, y)$ for any point $(x,y)$. Figure \ref{fig: d2c_angle} (c) and (d) respectively plot the histograms of $\Phi_{t}(x,y)$ on CIFAR-10 and CIFAR-100 at three distinct checkpoints $t$ and the the yellow curves in Figure \ref{fig: d2c_angle} (e) and (f) illustrate the evolution of $\mathbb{E}_{\mathcal{D}^*}\Phi_{t}(x,y)$. The results present an increasing trend of $\Phi_{t}(x,y)$, indicating a widening spread of the perturbation angles as PGD-AT proceed. This observation correlates well with the increasing trend of local dispersion.  Similar experimental results have been observed across other datasets. (see Appendix \ref{section: angle}). 

We conjecture that this wider spread of angles is due to the increasingly complex shape of the model's decision boundary along adversarial training. Notably the shape of the decision boundary has a substantial influence on the direction of adversarial perturbations. For instance,
adversarial perturbations against a linear classifiers are always aligned, namely, all perpendicular to the decision hyperplane. As the decision boundary bends to increasingly sharper angles and to more complex shapes, adversarial perturbation on points in a small neigbourhood of a robust anchor can vary greatly in directions.

We speculate that during the early stage of PGD-AT, the model's decision boundary quickly arrive at a smooth shape, causing adversarial perturbations to  exhibit a higher degree of alignment. This results in a smaller level of  $\mathbb{E}_{\mathcal{D}^*}\Phi_{t}(x,y)$ and  $\mathbb{E}_{\mathcal{D}^*}\tilde{\gamma}_t(x,y)$. However, as training progresses, the decision boundary evolves to a more complex shape, in order to fit or "memorize" the adversarially perturbed training data. This causing the subsequent adversarial perturbations to become less aligned and more dispersed, resulting in the rise of $\mathbb{E}_{\mathcal{D}^*}\Phi_{t}(x,y)$ and  $\mathbb{E}_{\mathcal{D}^*}\tilde{\gamma}_t(x,y)$.
Consequently, the increasing dispersion or spread of angles may serve as an indicative measure for the degree of irregularity, or the ``shape complexity'', of the decision boundary.

Our observation of the increasing dispersion and spreading angles of adversarial perturbations along PGD-AT is, to our best knowledge, a novel finding. This discovery may provide valuable insights into comprehending the dynamics of PGD-AT and the phenomenon of robust overfitting.

\section{Conclusion}

In this paper, we show that adversarial perturbation induced distribution in PGD-AT plays an important role in robust overfitting. In particular, we observe experimentally that the increasing generalization difficulty of the induced distribution along the PGD-AT trajectory is correlated with robust overfitting. Our theoretical analysis suggests that a key factor governing this difficulty is the local dispersion of the perturbation. Experimental results confirm that as PGD-AT proceeds, the perturbation becomes more dispersed, validating our theoretical results. Various additional insights are also presented.

This work points to a new direction in the search of explanations for robust overfitting. Remarkably, through this work, we demonstrate that the trajectory of PGD-AT plays an important role in robust overfitting. Studying the dynamics of adversarial training is arguably a promising approach to developing deeper understanding of this topic. In particular, we speculate that studying the effect of gradient-based parameter update may provide additional insight.

\bibliographystyle{unsrt}  
\bibliography{references}

\begin{thebibliography}{10}

\bibitem{szegedy2014intriguing}
Christian Szegedy, Wojciech Zaremba, Ilya Sutskever, Joan Bruna, Dumitru Erhan, Ian Goodfellow, and Rob Fergus.
\newblock Intriguing properties of neural networks, 2014.

\bibitem{goodfellow2015explaining}
Ian~J. Goodfellow, Jonathon Shlens, and Christian Szegedy.
\newblock Explaining and harnessing adversarial examples, 2015.

\bibitem{madry2019deep}
Aleksander Madry, Aleksandar Makelov, Ludwig Schmidt, Dimitris Tsipras, and Adrian Vladu.
\newblock Towards deep learning models resistant to adversarial attacks, 2019.

\bibitem{DBLP:journals/corr/abs-1901-08573}
Hongyang Zhang, Yaodong Yu, Jiantao Jiao, Eric~P. Xing, Laurent~El Ghaoui, and Michael~I. Jordan.
\newblock Theoretically principled trade-off between robustness and accuracy.
\newblock {\em CoRR}, abs/1901.08573, 2019.

\bibitem{DBLP:journals/corr/abs-2010-09670}
Francesco Croce, Maksym Andriushchenko, Vikash Sehwag, Nicolas Flammarion, Mung Chiang, Prateek Mittal, and Matthias Hein.
\newblock Robustbench: a standardized adversarial robustness benchmark.
\newblock {\em CoRR}, abs/2010.09670, 2020.

\bibitem{DBLP:journals/corr/abs-1802-00420}
Anish Athalye, Nicholas Carlini, and David~A. Wagner.
\newblock Obfuscated gradients give a false sense of security: Circumventing defenses to adversarial examples.
\newblock {\em CoRR}, abs/1802.00420, 2018.

\bibitem{dong2020benchmarking}
Yinpeng Dong, Qi-An Fu, Xiao Yang, Tianyu Pang, Hang Su, Zihao Xiao, and Jun Zhu.
\newblock Benchmarking adversarial robustness on image classification.
\newblock In {\em proceedings of the IEEE/CVF conference on computer vision and pattern recognition}, pages 321--331, 2020.

\bibitem{DBLP:journals/corr/abs-2002-11569}
Leslie Rice, Eric Wong, and J.~Zico Kolter.
\newblock Overfitting in adversarially robust deep learning.
\newblock {\em CoRR}, abs/2002.11569, 2020.

\bibitem{krizhevsky2009learning}
Alex Krizhevsky, Geoffrey Hinton, et~al.
\newblock Learning multiple layers of features from tiny images.
\newblock 2009.

\bibitem{DBLP:journals/corr/ZagoruykoK16}
Sergey Zagoruyko and Nikos Komodakis.
\newblock Wide residual networks.
\newblock {\em CoRR}, abs/1605.07146, 2016.

\bibitem{DBLP:journals/corr/abs-2004-05884}
Dongxian Wu, Yisen Wang, and Shutao Xia.
\newblock Revisiting loss landscape for adversarial robustness.
\newblock {\em CoRR}, abs/2004.05884, 2020.

\bibitem{DBLP:journals/corr/abs-2104-04448}
David Stutz, Matthias Hein, and Bernt Schiele.
\newblock Relating adversarially robust generalization to flat minima.
\newblock {\em CoRR}, abs/2104.04448, 2021.

\bibitem{chen2021robust}
Tianlong Chen, Zhenyu Zhang, Sijia Liu, Shiyu Chang, and Zhangyang Wang.
\newblock Robust overfitting may be mitigated by properly learned smoothening.
\newblock In {\em International Conference on Learning Representations}, 2021.

\bibitem{DBLP:journals/corr/abs-2102-07861}
Vasu Singla, Sahil Singla, David Jacobs, and Soheil Feizi.
\newblock Low curvature activations reduce overfitting in adversarial training.
\newblock {\em CoRR}, abs/2102.07861, 2021.

\bibitem{DBLP:journals/corr/abs-2110-03135}
Chengyu Dong, Liyuan Liu, and Jingbo Shang.
\newblock Double descent in adversarial training: An implicit label noise perspective.
\newblock {\em CoRR}, abs/2110.03135, 2021.

\bibitem{DBLP:journals/corr/abs-2106-01606}
Yinpeng Dong, Ke~Xu, Xiao Yang, Tianyu Pang, Zhijie Deng, Hang Su, and Jun Zhu.
\newblock Exploring memorization in adversarial training.
\newblock {\em CoRR}, abs/2106.01606, 2021.

\bibitem{yu2022understanding}
Chaojian Yu, Bo~Han, Li~Shen, Jun Yu, Chen Gong, Mingming Gong, and Tongliang Liu.
\newblock Understanding robust overfitting of adversarial training and beyond, 2022.

\bibitem{smoothness}
Sekitoshi Kanai, Masanori Yamada, Hiroshi Takahashi, Yuki Yamanaka, and Yasutoshi Ida.
\newblock Relationship between nonsmoothness in adversarial training, constraints of attacks, and flatness in the input space.
\newblock {\em IEEE Transactions on Neural Networks and Learning Systems}, 2023.
\newblock accepted.

\bibitem{hameed2022boundary}
Muhammad~Zaid Hameed and Beat Buesser.
\newblock Boundary adversarial examples against adversarial overfitting, 2022.

\bibitem{DBLP:journals/corr/abs-1804-11285}
Ludwig Schmidt, Shibani Santurkar, Dimitris Tsipras, Kunal Talwar, and Aleksander Madry.
\newblock Adversarially robust generalization requires more data.
\newblock {\em CoRR}, abs/1804.11285, 2018.

\bibitem{khim2019adversarial}
Justin Khim and Po-Ling Loh.
\newblock Adversarial risk bounds via function transformation, 2019.

\bibitem{DBLP:journals/corr/abs-1810-11914}
Dong Yin, Kannan Ramchandran, and Peter~L. Bartlett.
\newblock Rademacher complexity for adversarially robust generalization.
\newblock {\em CoRR}, abs/1810.11914, 2018.

\bibitem{DBLP:journals/corr/abs-2004-13617}
Pranjal Awasthi, Natalie Frank, and Mehryar Mohri.
\newblock Adversarial learning guarantees for linear hypotheses and neural networks.
\newblock {\em CoRR}, abs/2004.13617, 2020.

\bibitem{xiao2022adversarial}
Jiancong Xiao, Yanbo Fan, Ruoyu Sun, and Zhi-Quan Luo.
\newblock Adversarial rademacher complexity of deep neural networks, 2022.

\bibitem{DBLP:journals/corr/abs-1810-02180}
Idan Attias, Aryeh Kontorovich, and Yishay Mansour.
\newblock Improved generalization bounds for robust learning.
\newblock {\em CoRR}, abs/1810.02180, 2018.

\bibitem{DBLP:journals/corr/abs-1902-04217}
Omar Montasser, Steve Hanneke, and Nathan Srebro.
\newblock {VC} classes are adversarially robustly learnable, but only improperly.
\newblock {\em CoRR}, abs/1902.04217, 2019.

\bibitem{xing2021on}
Yue Xing, Qifan Song, and Guang Cheng.
\newblock On the algorithmic stability of adversarial training.
\newblock In A.~Beygelzimer, Y.~Dauphin, P.~Liang, and J.~Wortman Vaughan, editors, {\em Advances in Neural Information Processing Systems}, 2021.

\bibitem{xiao2022stability}
Jiancong Xiao, Yanbo Fan, Ruoyu Sun, Jue Wang, and Zhi-Quan Luo.
\newblock Stability analysis and generalization bounds of adversarial training, 2022.

\bibitem{cullina2018paclearning}
Daniel Cullina, Arjun~Nitin Bhagoji, and Prateek Mittal.
\newblock Pac-learning in the presence of evasion adversaries, 2018.

\bibitem{DBLP:journals/corr/abs-1906-05815}
Dimitrios~I. Diochnos, Saeed Mahloujifar, and Mohammad Mahmoody.
\newblock Lower bounds for adversarially robust {PAC} learning.
\newblock {\em CoRR}, abs/1906.05815, 2019.

\bibitem{li2022robust}
Binghui Li, Jikai Jin, Han Zhong, John~E. Hopcroft, and Liwei Wang.
\newblock Why robust generalization in deep learning is difficult: Perspective of expressive power, 2022.

\bibitem{DBLP:journals/corr/abs-1906-02931}
Yan Li, Ethan~X. Fang, Huan Xu, and Tuo Zhao.
\newblock Inductive bias of gradient descent based adversarial training on separable data.
\newblock {\em CoRR}, abs/1906.02931, 2019.

\bibitem{kuhn2019wasserstein}
Daniel Kuhn, Peyman~Mohajerin Esfahani, Viet~Anh Nguyen, and Soroosh Shafieezadeh-Abadeh.
\newblock Wasserstein distributionally robust optimization: Theory and applications in machine learning.
\newblock In {\em Operations research \& management science in the age of analytics}, pages 130--166. Informs, 2019.

\bibitem{sinha2020certifying}
Aman Sinha, Hongseok Namkoong, Riccardo Volpi, and John Duchi.
\newblock Certifying some distributional robustness with principled adversarial training, 2020.

\bibitem{staib2017distributionally}
Matthew Staib and Stefanie Jegelka.
\newblock Distributionally robust deep learning as a generalization of adversarial training.
\newblock In {\em NIPS workshop on Machine Learning and Computer Security}, volume~3, page~4, 2017.

\bibitem{bui2022unified}
Tuan~Anh Bui, Trung Le, Quan Tran, He~Zhao, and Dinh Phung.
\newblock A unified wasserstein distributional robustness framework for adversarial training.
\newblock {\em arXiv preprint arXiv:2202.13437}, 2022.

\bibitem{bennouna2023certified}
Amine Bennouna, Ryan Lucas, and Bart Van~Parys.
\newblock Certified robust neural networks: Generalization and corruption resistance.
\newblock {\em arXiv preprint arXiv:2303.02251}, 2023.

\bibitem{bai2021recent}
Tao Bai, Jinqi Luo, Jun Zhao, Bihan Wen, and Qian Wang.
\newblock Recent advances in adversarial training for adversarial robustness, 2021.

\bibitem{qian2022survey}
Zhuang Qian, Kaizhu Huang, Qiu-Feng Wang, and Xu-Yao Zhang.
\newblock A survey of robust adversarial training in pattern recognition: Fundamental, theory, and methodologies, 2022.

\bibitem{lecun1998gradient}
Yann LeCun, L{\'e}on Bottou, Yoshua Bengio, and Patrick Haffner.
\newblock Gradient-based learning applied to document recognition.
\newblock {\em Proceedings of the IEEE}, 86(11):2278--2324, 1998.

\bibitem{russakovsky2015imagenet}
Olga Russakovsky, Jia Deng, Hao Su, Jonathan Krause, Sanjeev Satheesh, Sean Ma, Zhiheng Huang, Andrej Karpathy, Aditya Khosla, Michael Bernstein, et~al.
\newblock Imagenet large scale visual recognition challenge.
\newblock {\em International journal of computer vision}, 115:211--252, 2015.

\bibitem{tsipras2019robustness}
Dimitris Tsipras, Shibani Santurkar, Logan Engstrom, Alexander Turner, and Aleksander Madry.
\newblock Robustness may be at odds with accuracy, 2019.

\bibitem{DBLP:journals/corr/HeZR016}
Kaiming He, Xiangyu Zhang, Shaoqing Ren, and Jian Sun.
\newblock Identity mappings in deep residual networks.
\newblock {\em CoRR}, abs/1603.05027, 2016.

\bibitem{DBLP:journals/corr/abs-1710-10766}
Yang Song, Taesup Kim, Sebastian Nowozin, Stefano Ermon, and Nate Kushman.
\newblock Pixeldefend: Leveraging generative models to understand and defend against adversarial examples.
\newblock {\em CoRR}, abs/1710.10766, 2017.

\bibitem{gilmer2018adversarial}
Justin Gilmer, Luke Metz, Fartash Faghri, Samuel~S. Schoenholz, Maithra Raghu, Martin Wattenberg, and Ian Goodfellow.
\newblock Adversarial spheres, 2018.

\bibitem{raghunathan2019adversarial}
Aditi Raghunathan, Sang~Michael Xie, Fanny Yang, John~C. Duchi, and Percy Liang.
\newblock Adversarial training can hurt generalization, 2019.

\bibitem{khoury2018geometry}
Marc Khoury and Dylan Hadfield-Menell.
\newblock On the geometry of adversarial examples, 2018.

\bibitem{stutz2019disentangling}
David Stutz, Matthias Hein, and Bernt Schiele.
\newblock Disentangling adversarial robustness and generalization, 2019.

\bibitem{epsDeltaRobust:2023}
Ambar Pal, Jeremias Sulam, and René Vidal.
\newblock Adversarial examples mightbe avoidable: The role of data concentration in adversarial robustness.
\newblock In {\em NeurIPS 2024}, 2023.

\bibitem{DBLP:journals/corr/abs-2003-02460}
Yao{-}Yuan Yang, Cyrus Rashtchian, Hongyang Zhang, Ruslan Salakhutdinov, and Kamalika Chaudhuri.
\newblock Adversarial robustness through local lipschitzness.
\newblock {\em CoRR}, abs/2003.02460, 2020.

\end{thebibliography}

\newpage
\appendix

\section{Detailed Experimental setup}
\label{section: setup}

Our Reduced ImageNet is made by aggregating several semantically similar subsets of the original ImageNet, resulting in a total of 66594 images. This dataset is then partitioned into a training set containing 5,000 images per class and a testing set containing approximately 1,000 images per class. Compared to the restricted ImageNet in \cite{russakovsky2015imagenet}, our dataset has a more balanced sample size across each classes. Table \ref{Table: resimg} illustrates the specific classes from the original ImageNet that have been aggregated in our dataset.
\begin{table}[!h]\small
\centering
\setlength{\tabcolsep}{0.5mm}{
\begin{tabular}{lcc}  
\toprule
    & Classes in the reduced ImageNet \quad  & Classes in ImageNet  \\ 
\midrule
      & "dog"   & 86 to 90  \\
      & "cat"  & (8,10,55,95,174) \\
      & "truck"  & 279 to 283  \\
       & "car"  & 272 to 276   \\
        & "beetles" & 623 to 627   \\
       &"turtle"  &458 to 462  \\
       &"crab"   &612 to 616   \\
      & "fish" &450 to 454   \\
      &"snake"  & 477 to 481 \\
      & "spider" & 604 to 608 \\
\bottomrule
\end{tabular}}
\caption{The left column presents the classes within our reduced ImageNet dataset, with each class being an aggregation of the corresponding classes from the full-scale ImageNet dataset, as depicted in the right column.}
\label{Table: resimg}
\end{table}

For PGD-AT, the settings on different datasets are summarized in Table \ref{Table: adv}. Data augmentation is performed on 
these datasets except for MNIST during the training. For CIFAR-10 and CIFAR-100 we follow the data augmentation setting in  \cite{DBLP:journals/corr/abs-2002-11569}. For our reduced ImageNet, we adopt the same data augmentation scheme that is used on the restricted ImageNet in \cite{DBLP:journals/corr/abs-2003-02460}. 

\begin{table}[!h]\small
\centering
\setlength{\tabcolsep}{0.5mm}{
\begin{tabular}{lcccc}  
\toprule
                   &MNIST &CIFAR-10 &CIFAR-100 &Reduced ImageNet \\ 
\midrule
model    & small CNN  &PRN18\&WRN-34 &WRN-34  & PRN-50  \\
optimizer            & Adam  & SGD     & SGD & SGD \\
weight deacy         & None  & $5\times10^{-4}$ & $5\times10^{-4}$ & None \\
batch size           & 128  & 128  &128 &128 \\
$\epsilon$           & 0.3  & 8/255  &8/255 &4/255 \\
$\lambda$            & 0.01  &2/255 &2/255 &0.9/255 \\
number of PGD        & 40  & 10  &10  &5 \\
\bottomrule
\end{tabular}}
\caption{Settings in PGD-AT across different datasets}
\label{Table: adv}
\end{table}

For the IDEs on each datasets, the settings are outlined in Table \ref{Table: IDE}. It is important to note that for each of the individual IDEs that is conducted on the same dataset, we maintain consistent training settings. This includes using the same model architecture with identical model size and the same level of regularization. This ensures a fair comparison of the IDE results obtained from the same dataset. Furthermore, the model is trained to achieve zero training error in all the IDEs, excluding the situation that the degeneration in model performance could be attributed to inadequate training procedures.

\begin{table}[!h]\small
\centering
\setlength{\tabcolsep}{0.5mm}{
\begin{tabular}{lcccc}  
\toprule
                   &MNIST &CIFAR-10 &CIFAR-100 &Reduced ImageNet \\ 
\midrule
model    & small CNN  & WRN-34 &WRN-34  & PRN-50  \\
optimizer            & Adam   & SGD    & SGD & SGD \\
weight deacy         & None  & $5\times10^{-4}$ & $5\times10^{-4}$ & $5\times10^{-4}$ \\
batch size           & 128  &128 &128 &128 \\
\bottomrule
\end{tabular}}
\caption{Settings in the IDE across different datasets}
\label{Table: IDE}
\end{table}

\section{Proof of the theorem}
\label{section: proof}
We follow the notations introduced in the main text. Recall that $(x,y)$ denotes an instance drawn from the distribution ${\cal D}^{*}$ and $(v,y)$ denotes an instance drawn from the associate induced distribution $\tilde{\mathcal{D}}_{t}$. For shorter notations, we will denote $z:=(x,y)$, $u:=(v,y)$ and $f(u):=f(v,y)$ and write $\mathcal{Q}_{x,y,\theta_t}$ as $\mathcal{Q}_{z, \theta_t}$. 

Denote by $g(u_1\cdots u_m):=\left|\frac{1}{m}\sum\limits_{i=1}^{m}f(u_{i})-\mathbb{E}f(u)\right|$. We have for any $1\le j\le m$

\begin{align}
&\sup\limits_{u_1,\cdots ,u_m, u_j'}\left|g(u_1,\cdots ,u_m)-g(u_1,\cdots ,u_j', u_{j+1},\cdots u_m)\right|\\
= & \sup\limits_{u_1,\cdots ,u_m, u_j'}\left|\left|\frac{1}{m}\sum\limits_{i=1}^{m}f(u_{i})-\mathbb{E}f(u)\right|-\left|\frac{1}{m}\left(\sum\limits_{i=1,i\ne j}^{m}f(u_{i}) +f(u_{j}')\right) -\mathbb{E}_{u}f(u)\right|\right|\\
\le & \sup\limits_{u_1,\cdots ,u_m, u_j'}\left|\frac{1}{m}\sum\limits_{i=1}^{m}f(u_{i})-\mathbb{E}_{u}f(u)-\frac{1}{m}\left(\sum\limits_{i=1,i\ne j}^{m}f(u_{i}) +f(u_{j}')\right) + \mathbb{E}_{u}f(u)\right| \label{eq 1} \\ 
=& \sup\limits_{u_j, u_j'}\frac{1}{m}\left|f(u_j)-f(u_j')\right|\\
\le & \frac{1}{m}\sup\limits_{u_j}\left|f(u_j)\right| + \frac{1}{m}\sup\limits_{u_j'}\left|f(u_j')\right| \label{eq 2} \\
\le & \frac{2B}{m} \label{eq 3}
\end{align}

where the inequality (\ref{eq 1}) follows from the inverse triangle inequality. The inequality (\ref{eq 2}) and (\ref{eq 3}) make use of the triangle inequality and the boundedness condition of $f$.

With the result derived above, by McDiarmid inequality, we have for all $\mu>0$
$$
{\rm Pr}\left[g(u_1\cdots u_m)-\mathbb{E}_{U}g(u_1\cdots u_m)\ge \mu  \right]\le \exp\left(\frac{-m\mu^{2}}{B}\right)
$$
where we use $U:=(u_1,\cdots,u_m)$. This is equivalent to saying that with probability $1-\tau$, we have
\begin{equation}
    g(u_1\cdots u_m)\le \mathbb{E}_{U}g(u_1\cdots u_m)+2B\sqrt{\frac{\log\frac{1}{\tau}}{2m}}
    \label{eq: mcdiarmid}
\end{equation}
Given this, the following parts aim at constructing an upper bound for the term $\mathbb{E}_{U}g(u_1\cdots u_m)$.

 For shorter notations, let $Z:=(z_1,\cdots, z_m)$, $\Gamma:=(\rho_1,\cdots, \rho_m)$ and $F(Z, \Gamma):=\frac{1}{m}\sum\limits_{i=1}^{m}f(\mathcal{Q}_{z_i, \theta_t}(x_i+\rho_i), y_i)$. We have
\begin{align}
    &\mathbb{E}_{U}g(u_1\cdots u_m)\\
=& \mathbb{E}_{U}\left| \frac{1}{m}\sum\limits_{i=1}^{m}f(u_{i})-\mathbb{E}f(u)\right|\\
=& \mathbb{E}_{U}\left| \frac{1}{m}\sum\limits_{i=1}^{m}f(u_{i})-\mathbb{E}_{\hat{U}}\left[\frac{1}{m}\sum\limits_{i=1}^{m}f(\hat{u}_i)\right]\right|\\
\le & \mathbb{E}_{U}\mathbb{E}_{\hat{U}}\left| \frac{1}{m}\sum\limits_{i=1}^{m}f(u_{i})-\frac{1}{m}\sum\limits_{i=1}^{m}f(\hat{u}_i)\right| \label{eq 4} \\
=& \mathbb{E}_{Z}\mathbb{E}_{\Gamma}\mathbb{E}_{\hat{Z}}\mathbb{E}_{\hat{\Gamma
}}\left|\frac{1}{m}\sum\limits_{i=1}^{m}f(\mathcal{Q}_{z_i, \theta_t}(x_i+\rho_i), y_i)-\frac{1}{m}\sum\limits_{i=1}^{m}f(\mathcal{Q}_{\hat{z}_i, \theta_t}(\hat{x}_i+\hat{\rho}_i), \hat{y}_i) \right|\\
=& \mathbb{E}_{Z}\mathbb{E}_{\Gamma}\mathbb{E}_{\hat{Z}}\mathbb{E}_{\hat{\Gamma
}}\left|F(Z, \Gamma) -
\mathbb{E}_{\bar{\Gamma}}F(Z,\bar{\Gamma})+ \mathbb{E}_{\bar{\Gamma}}F(Z,\bar{\Gamma}) -F(\hat{Z}, \hat{\Gamma})+\mathbb{E}_{\tilde{\Gamma}}F(\hat{Z}, \tilde{\Gamma})-\mathbb{E}_{\tilde{\Gamma}}F(\hat{Z}, \tilde{\Gamma}) \right|\\
\le & \mathbb{E}_{Z}\mathbb{E}_{\Gamma}\left|F(Z, \Gamma) -
\mathbb{E}_{\bar{\Gamma}}F(Z,\bar{\Gamma}) \right| + \mathbb{E}_{\hat{Z}}\mathbb{E}_{\hat{\Gamma
}}\left|F(\hat{Z}, \hat{\Gamma})- \mathbb{E}_{\tilde{\Gamma}}F(\hat{Z}, \tilde{\Gamma}) \right| + \mathbb{E}_{Z}\mathbb{E}_{\hat{Z}}\left|\mathbb{E}_{\bar{\Gamma}}F(Z,\bar{\Gamma})-\mathbb{E}_{\tilde{\Gamma}}F(\hat{Z}, \tilde{\Gamma})\right| \label{eq 5} \\
=&\underbrace{2\mathbb{E}_{Z}\mathbb{E}_{\Gamma}\left|F(Z, \Gamma) -
\mathbb{E}_{\bar{\Gamma}}F(Z,\bar{\Gamma}) \right|}_{\textcircled{1}} +\underbrace{\mathbb{E}_{Z}\mathbb{E}_{\hat{Z}}\left|\mathbb{E}_{\bar{\Gamma}}F(Z,\bar{\Gamma})-\mathbb{E}_{\tilde{\Gamma}}F(\hat{Z}, \tilde{\Gamma})\right|}_{\textcircled{2}} \label{eq final}
\end{align}
where (\ref{eq 4}) follows from Jensen's inequality and (\ref{eq 5}) is by the triangle inequality. We now individually construct upper bounds for the term $\textcircled{1}$ and $\textcircled{2}$.

For the term $\textcircled{1}$, we have
\begin{align}
 &2\mathbb{E}_{Z}\mathbb{E}_{\Gamma}\left|F(Z, \Gamma) -
\mathbb{E}_{\bar{\Gamma}}F(Z,\bar{\Gamma}) \right|\\
\le & 2\mathbb{E}_{Z}\mathbb{E}_{\Gamma}\mathbb{E}_{\bar{\Gamma}}\left|F(Z, \Gamma)-F(Z,\bar{\Gamma}) \right| \label{eq 5}\\
=& 2\mathbb{E}_{Z}\mathbb{E}_{\Gamma}\mathbb{E}_{\bar{\Gamma}}\left|\frac{1}{m}\sum\limits_{i=1}^{m}f(\mathcal{Q}_{z_i,\theta_t}(x_i+\rho_i), y_i)-\frac{1}{m}\sum\limits_{i=1}^{m}f(\mathcal{Q}_{z_i,\theta_t}(x_i+\bar{\rho}_i), y_i) \right|\\
=&\frac{2}{m}\mathbb{E}_{Z}\mathbb{E}_{\Gamma}\mathbb{E}_{\bar{\Gamma}}\mathbb{E}_{ \Sigma}\left|\sum\limits_{i=1}^{m}\sigma_i \left(f(\mathcal{Q}_{z_i,\theta_t}(x_i+\rho_i), y_i)-f(\mathcal{Q}_{z_i,\theta_t}(x_i+\bar{\rho}_i), y_i)\right) \right| \label{eq 6} \\
\le & \frac{2}{m}\mathbb{E}_{Z}\mathbb{E}_{\Gamma}\mathbb{E}_{\bar{\Gamma}}\sqrt{\sum\limits_{i=1}^{m}\left|f(\mathcal{Q}_{z_i,\theta_t}(x_i+\rho_i), y_i)-f(\mathcal{Q}_{z_i,\theta_t}(x_i+\bar{\rho}_i), y_i)\right|^{2}} \label{eq 7}\\
\le & \frac{2}{m}\mathbb{E}_{Z}\mathbb{E}_{\Gamma}\mathbb{E}_{\bar{\Gamma}}\sqrt{\sum\limits_{i=1}^{m}\beta^{2}\|\mathcal{Q}_{z_i,\theta_t}(x_i+\rho_i)-\mathcal{Q}_{z_i,\theta_t}(x_i+\bar{\rho}_i)\|^{2}} \label{eq 8}\\
\le & \frac{2\beta}{m}\mathbb{E}_{Z}\sqrt{\mathbb{E}_{\Gamma}\mathbb{E}_{\bar{\Gamma}}
\left[\sum\limits_{i=1}^{m}\|\mathcal{Q}_{z_i,\theta_t}(x_i+\rho_i)-\mathcal{Q}_{z_i,\theta_t}(x_i+\bar{\rho}_i)\|^{2}\right]}  \label{eq 9} \\
= & \frac{2\beta}{m}\mathbb{E}_{Z}\sqrt{
\sum\limits_{i=1}^{m}\mathbb{E}_{\rho}\mathbb{E}_{\bar{\rho}}\|\mathcal{Q}_{z_i,\theta_t}(x_i+\rho)-\mathcal{Q}_{z_i,\theta_t}(x_i+\bar{\rho})\|^{2}}  \\
=& \frac{2\beta}{m}\mathbb{E}_{Z}\sqrt{\sum\limits_{i=1}^{m}\gamma_t(x_i,y_i)} \label{eq 10-0} \\
\le & \frac{2\beta}{m}\sqrt{\mathbb{E}_{Z}\left[\sum\limits_{i=1}^{m}\gamma_t(x_i,y_i)\right]}  \label{eq 10} \\
= &\frac{2\beta}{m}\sqrt{\sum\limits_{i=1}^{m}\mathbb{E}_{z_i}\gamma_t(x_i,y_i)}  \label{eq 11} \\
=&  \frac{2\beta}{\sqrt{m}}\sqrt{\mathbb{E}_{z}\gamma_t(x,y)}  \label{eq 12}
\end{align}
Again, we apply Jensen's inequality to get (\ref{eq 5}). In (\ref{eq 6}), we introduce Rademacher variables $\Sigma:=(\sigma_1,\cdots,\sigma_m)$ (i.e., each random variable $\sigma_i$ takes values in $\{-1,+1\}$ independently with equal probability 0.5). The Rademacher variables introduces a random exchange of the corresponding difference term. Since $\Gamma$ and $\hat{\Gamma}$ are independently sampled from the same distribution, such a swap gives an equally likely configuration. Therefore, the equality (\ref{eq 6}) holds. The inequality (\ref{eq 7}) is given by Khintchine's inequality. The inequality (\ref{eq 8}) makes use of the lipschitz condition of $f$. (\ref{eq 9}) is derived from Jensen's inequality and due to that square root is a concave function. (\ref{eq 10-0}) is by the definition of the local dispersion of $\mathcal{Q}_{\theta_t}$. Again, we apply Jensen's inequality to obtain (\ref{eq 10}). Equation (\ref{eq 11}) and (\ref{eq 12}) follow from the settings that each $z_i=(x_i,y_i)$ is i.i.d.

For the term $\textcircled{2}$, we have
\begin{align}
&\mathbb{E}_{Z}\mathbb{E}_{\hat{Z}}\left|\mathbb{E}_{\bar{\Gamma}}F(Z,\bar{\Gamma})-\mathbb{E}_{\tilde{\Gamma}}F(\hat{Z}, \tilde{\Gamma})\right| \\
=& \mathbb{E}_{Z}\mathbb{E}_{\hat{Z}}\left| \mathbb{E}_{\bar{\Gamma}} \left[\frac{1}{m}\sum\limits_{i=1}^{m}f(\mathcal{Q}_{z_i, \theta_t}(x_i+\bar{\rho}_i), y_i)\right]-   \mathbb{E}_{\tilde{\Gamma}} \left[\frac{1}{m}\sum\limits_{i=1}^{m}f(\mathcal{Q}_{z_i, \theta_t}(\hat{x}_i+\tilde{\rho}_i), \hat{y}_i)\right]  \right|\\
=& \mathbb{E}_{Z}\mathbb{E}_{\hat{Z}}\left| \frac{1}{m}\sum\limits_{i=1}^{m} \mathbb{E}_{\bar{\rho}_{i}} \left[f(\mathcal{Q}_{z_i, \theta_t}(x_i+\bar{\rho}_i), y_i)\right]-   \frac{1}{m}\sum\limits_{i=1}^{m}\mathbb{E}_{\tilde{\rho}_{i}} \left[f(\mathcal{Q}_{z_i, \theta_t}(\hat{x}_i+\tilde{\rho}_i), \hat{y}_i)\right]  \right|\label{eq 13}\\
=& \mathbb{E}_{Z}\mathbb{E}_{\hat{Z}}\left| \frac{1}{m}\sum\limits_{i=1}^{m} \mathbb{E}_{\rho} \left[f(\mathcal{Q}_{z_i, \theta_t}(x_i+\rho), y_i)\right]-   \frac{1}{m}\sum\limits_{i=1}^{m}\mathbb{E}_{\rho} \left[f(\mathcal{Q}_{z_i, \theta_t}(\hat{x}_i+\rho), \hat{y}_i)\right]  \right|\label{eq 14}\\
=& \frac{1}{m} \mathbb{E}_{Z}\mathbb{E}_{\hat{Z}}\mathbb{E}_{\Sigma}\left|\sum\limits_{i=1}^{m} \sigma_{i}\left(\mathbb{E}_{\rho} \left[f(\mathcal{Q}_{z_i, \theta_t}(x_i+\rho), y_i)\right]-\mathbb{E}_{\rho} \left[f(\mathcal{Q}_{z_i, \theta_t}(\hat{x}_i+\rho), \hat{y}_i)\right]\right)  \right|\label{eq 15}\\
\le &  \frac{1}{m} \mathbb{E}_{Z}\mathbb{E}_{\hat{Z}}\sqrt{\sum\limits_{i=1}^{m} \left|\left(\mathbb{E}_{\rho} \left[f(\mathcal{Q}_{z_i, \theta_t}(x_i+\rho), y_i)\right]-\mathbb{E}_{\rho} \left[f(\mathcal{Q}_{z_i, \theta_t}(\hat{x}_i+\rho), \hat{y}_i)\right]\right)  \right|^{2}}\label{eq 16}
\end{align}
where equation (\ref{eq 13}) and (\ref{eq 14}) are due to each $\hat{\rho}_{i}$ and $\tilde{\rho}_{i}$ is i.i.d. Again, we introduce Rademacher variables at (\ref{eq 15}) and apply Khintchine's inequality to get (\ref{eq 16}). For the term $\left|\left(\mathbb{E}_{\rho} \left[f(\mathcal{Q}_{z_i, \theta_t}(x_i+\rho), y_i)\right]-\mathbb{E}_{\rho} \left[f(\mathcal{Q}_{z_i, \theta_t}(\hat{x}_i+\rho), \hat{y}_i)\right]\right) \right|^{2}$, we have
\begin{align}
    &\left| \mathbb{E}_{\rho}f(\mathcal{Q}_{z_i, \theta_t}(x_i+\rho), y_i)-\mathbb{E}_{\rho}f(\mathcal{Q}_{z_i, \theta_t}(\hat{x}_i+\rho), \hat{y}_i) \right|^{2}\\
\le & (\left| \mathbb{E}_{\rho}f(\mathcal{Q}_{z_i, \theta_t}(x_i+\rho), y_i)\right|+ \left|\mathbb{E}_{\rho} f(\mathcal{Q}_{z_i, \theta_t}(\hat{x}_i+\rho), \hat{y}_i)  \right|)^{2}\\
\le & 2\left| \mathbb{E}_{\rho}f(\mathcal{Q}_{z_i, \theta_t}(x_i+\rho), y_i)\right|^{2}+2\left|\mathbb{E}_{\rho} f(\mathcal{Q}_{z_i, \theta_t}(\hat{x}_i+\rho), \hat{y}_i)  \right|^{2} \label{eq 17}
\end{align}
where inequality (\ref{eq 17}) is derived by the inequality $(a+b)^{2}\le 2(a^2+b^2)$. Returning to (\ref{eq 16}), we then have
\begin{align}
&\frac{1}{m} \mathbb{E}_{Z}\mathbb{E}_{\hat{Z}}\sqrt{\sum\limits_{i=1}^{m} \left|\left(\mathbb{E}_{\rho} \left[f(\mathcal{Q}_{z_i, \theta_t}(x_i+\rho), y_i)\right]-\mathbb{E}_{\rho} \left[f(\mathcal{Q}_{z_i, \theta_t}(\hat{x}_i+\rho), \hat{y}_i)\right]\right)  \right|^{2}} \notag \\
\le & \frac{1}{m} \mathbb{E}_{Z}\mathbb{E}_{\hat{Z}}\sqrt{\sum\limits_{i=1}^{m} 2\left| \mathbb{E}_{\rho}f(\mathcal{Q}_{z_i, \theta_t}(x_i+\rho), y_i)\right|^{2}+\sum\limits_{i=1}^{m}2\left|\mathbb{E}_{\rho} f(\mathcal{Q}_{z_i, \theta_t}(\hat{x}_i+\rho), \hat{y}_i)  \right|^{2}}\\
\le & \frac{1}{m} \sqrt{\mathbb{E}_{Z}\mathbb{E}_{\hat{Z}}\left[\sum\limits_{i=1}^{m} 2\left| \mathbb{E}_{\rho}f(\mathcal{Q}_{z_i, \theta_t}(x_i+\rho), y_i)\right|^{2}+\sum\limits_{i=1}^{m}2\left|\mathbb{E}_{\rho} f(\mathcal{Q}_{z_i, \theta_t}(\hat{x}_i+\rho), \hat{y}_i)  \right|^{2}\right]}\\
= & \frac{1}{m} \sqrt{\sum\limits_{i=1}^{m} 2\mathbb{E}_{z_i}\left| \mathbb{E}_{\rho}f(\mathcal{Q}_{z_i, \theta_t}(x_i+\rho), y_i)\right|^{2}+\sum\limits_{i=1}^{m}2\mathbb{E}_{\hat{z}_i}\left|\mathbb{E}_{\rho} f(\mathcal{Q}_{z_i, \theta_t}(\hat{x}_i+\rho), \hat{y}_i)  \right|^{2}}\\
= &\frac{2}{\sqrt{m}}\sqrt{\mathbb{E}_{z}\left| \mathbb{E}_{\rho}f(\mathcal{Q}_{z, \theta_t}(x+\rho), y)\right|^{2}}\\
\le & \frac{2}{\sqrt{m}}\sqrt{\mathbb{E}_{z}(\left| \mathbb{E}_{\rho}f(\mathcal{Q}_{z, \theta_t}(x+\rho), y)-f(x, y)\right|+\left|f(x, y)\right|)^{2}}\label{eq 18}\\
\le & \frac{2}{\sqrt{m}}\sqrt{\mathbb{E}_{z}(\mathbb{E}_{\rho}\left|f(\mathcal{Q}_{z, \theta_t}(x+\rho), y)-f(x, y)\right|+B)^{2}}\label{eq 19}\\
\le & \frac{2}{\sqrt{m}}\sqrt{\mathbb{E}_{z}(\mathbb{E}_{\rho}\beta\|\mathcal{Q}_{z, \theta_t}(x+\rho)-x\|_2+B)^{2}}\label{eq 20}\\
\le & \frac{2(\beta\sqrt{d}\epsilon+B)}{\sqrt{m}} \label{eq 21}
\end{align}

The inequalities (\ref{eq 18})-(\ref{eq 20}) respectively make use of the triangle inequality, Jensen's inequality, the boundedness and lipschitz condition of $f$. The final line is due to that with $\|\mathcal{Q}_{z, \theta_t}(x+\rho)-x\|_{\infty}\le \epsilon$ we have $\|\mathcal{Q}_{z, \theta_t}(x+\rho)-x\|_{2}\le \sqrt{d}\epsilon$. This gives the final result
$$\mathbb{E}_{U}g(u_1\cdots u_m)\le \frac{2\beta}{\sqrt{m}}\sqrt{\mathbb{E}_{z}\gamma(x,y)} +  \frac{2(\beta\sqrt{d}\epsilon+B)}{\sqrt{m}} $$
Plugging back to (\ref{eq: mcdiarmid}), we derive the bound in the Theorem. \hfill $\Box$

\section{Local dispersion results across other datasets}
\label{section: disper}

\begin{figure}[!htbp]
    \centering
    
    \subfigure[Reduced ImageNet]{\includegraphics[width=0.32\textwidth]{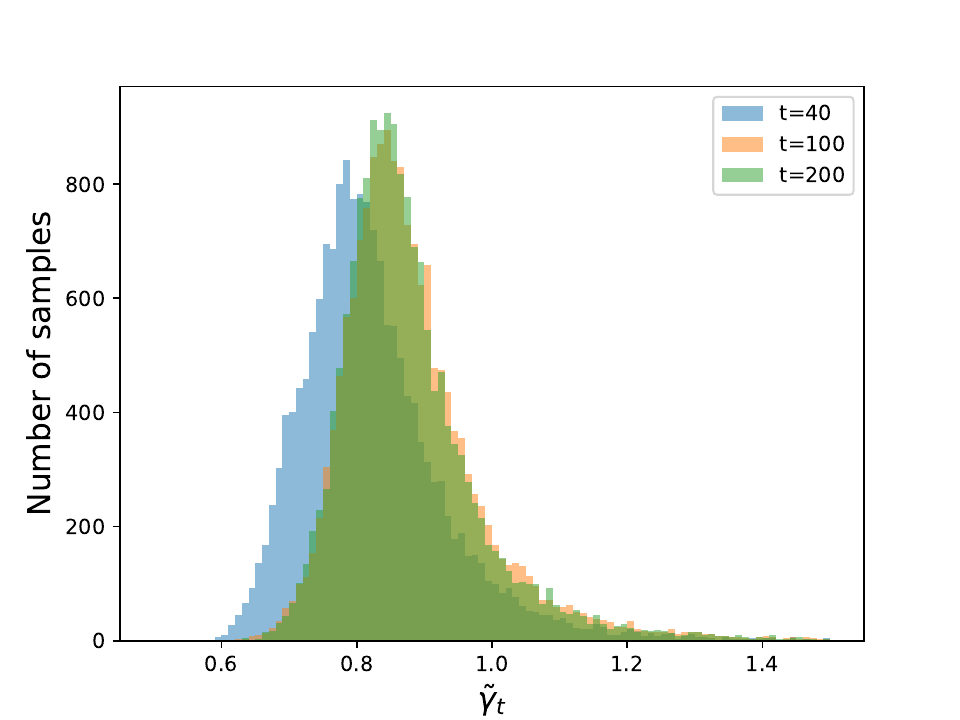}}
    \subfigure[MNIST]{\includegraphics[width=0.32\textwidth]{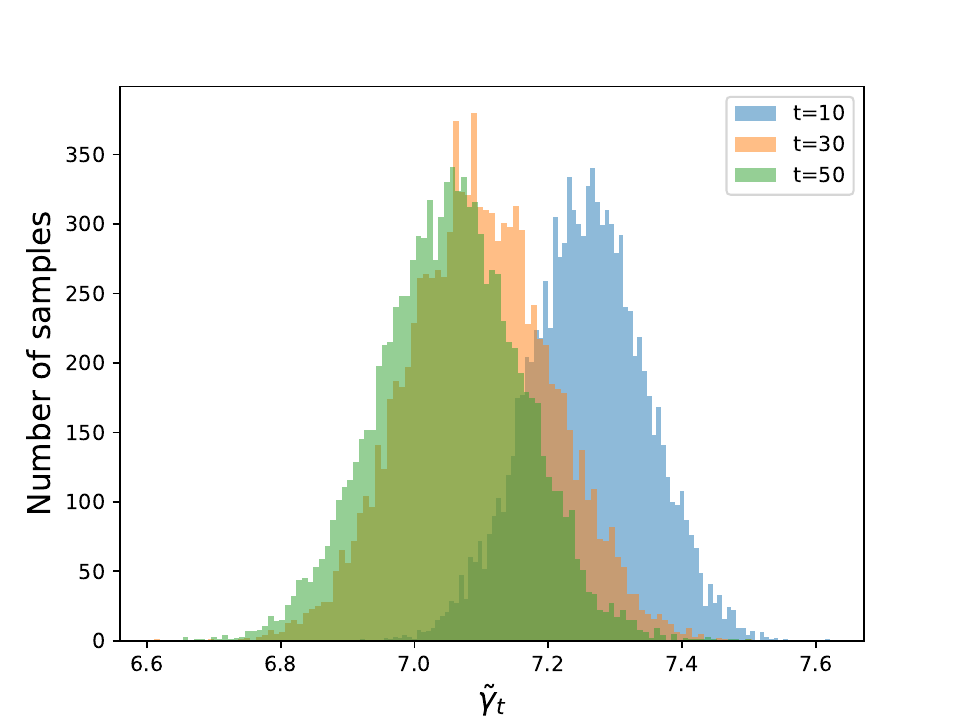}}
 
      \caption{histograms of $\tilde{\gamma}_t$ on the Reduced ImageNet and MNIST testing set. On the Reduced ImageNet, the mode of the histogram shifts towards a larger number, indicating the level of $\tilde{\gamma}_t$ increases along PGD-AT. By sharp contrast, on MNIST, the mode of the histogram shifts toward a smaller value. This behaviour matches the IDE results and the generalization bound derived in the Theorem.}
    \label{fig: disper_hist_extra}
\end{figure}

\begin{figure}[!htbp]
    \centering
    
    \subfigure[Reduced ImageNet]{\includegraphics[width=0.32\textwidth]{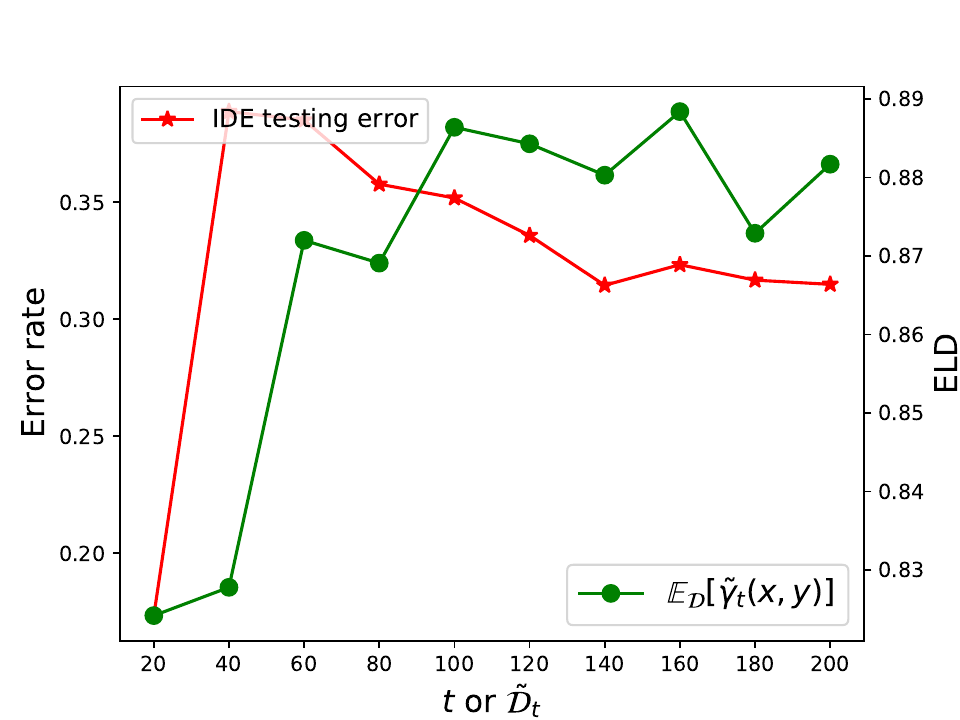}}
    \subfigure[MNIST]{\includegraphics[width=0.32\textwidth]{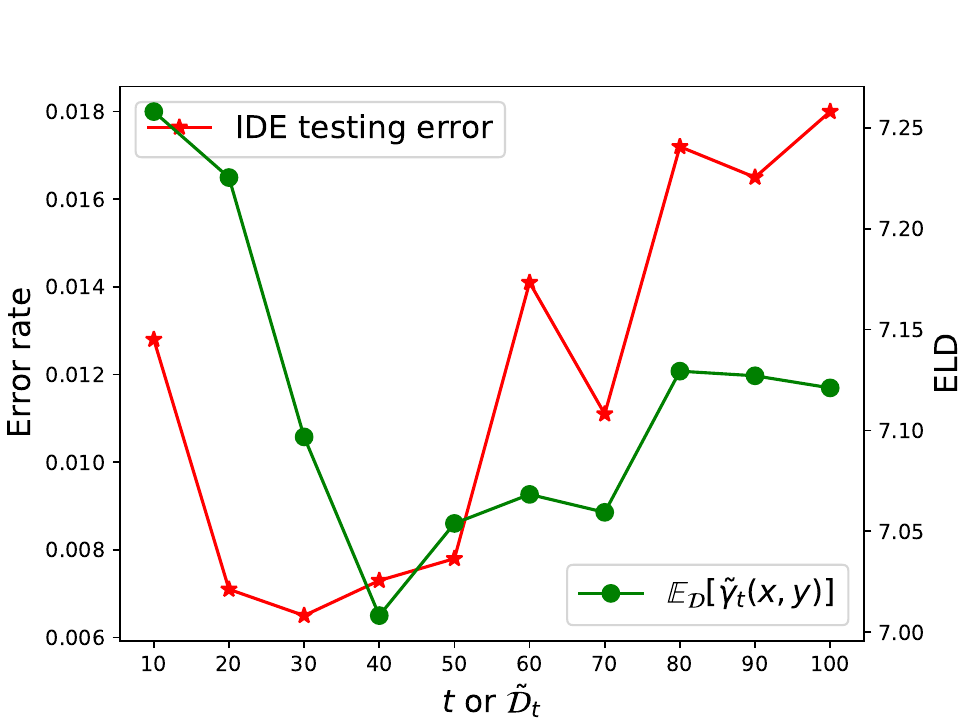}}
 
      \caption{ELD (green curves) evaluated on the Reduced ImageNet and MNIST testing set and the corresponding IDE results (red curves). In each figure, the green and red curves are tightly correlated. This provides additional support to the conclusion in the Theorem.}
\label{fig: disper_line_extra}
\end{figure}

\section{Additional results for section 6}
\label{section: angle}

\begin{figure}[!htbp]
    \centering
    \subfigure[Reduced ImageNet]{\includegraphics[width=0.32\textwidth]{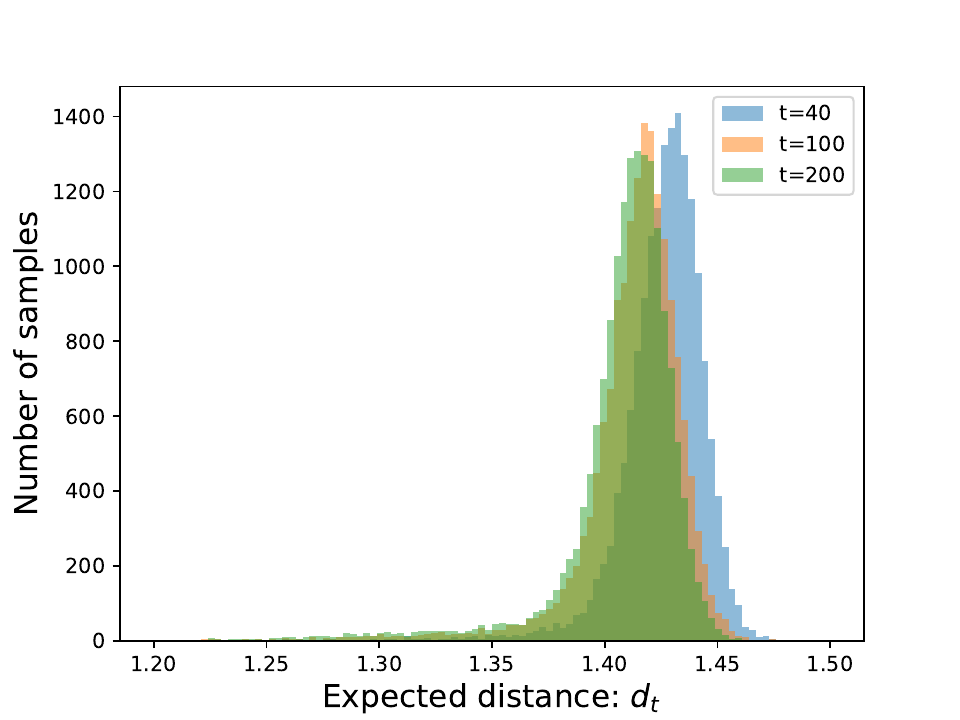}}
    \subfigure[MNIST]{\includegraphics[width=0.32\textwidth]{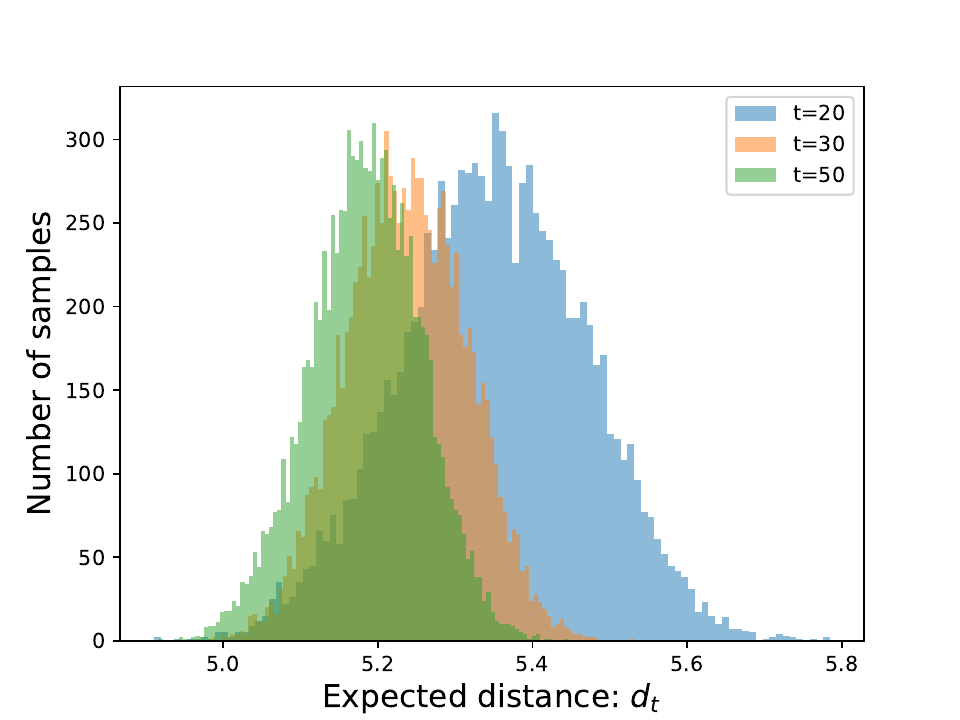}}
 
      \caption{histograms of $d_t$ on the Reduced ImageNet and MNIST testing set. The reduction in the level of $d_t$ along PGD-AT is shown in the figures.}
\label{fig: d2c_extra}
\end{figure}

\begin{figure}[!htbp]
    \centering
    
    \subfigure[Reduced ImageNet]{\includegraphics[width=0.32\textwidth]{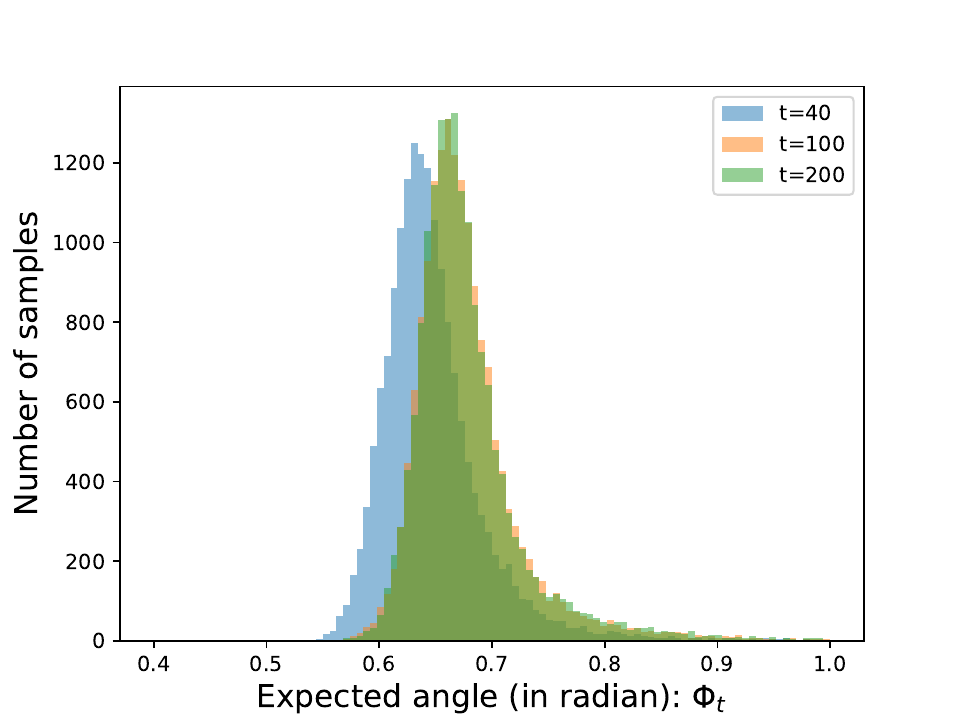}}
    \subfigure[MNIST]{\includegraphics[width=0.32\textwidth]{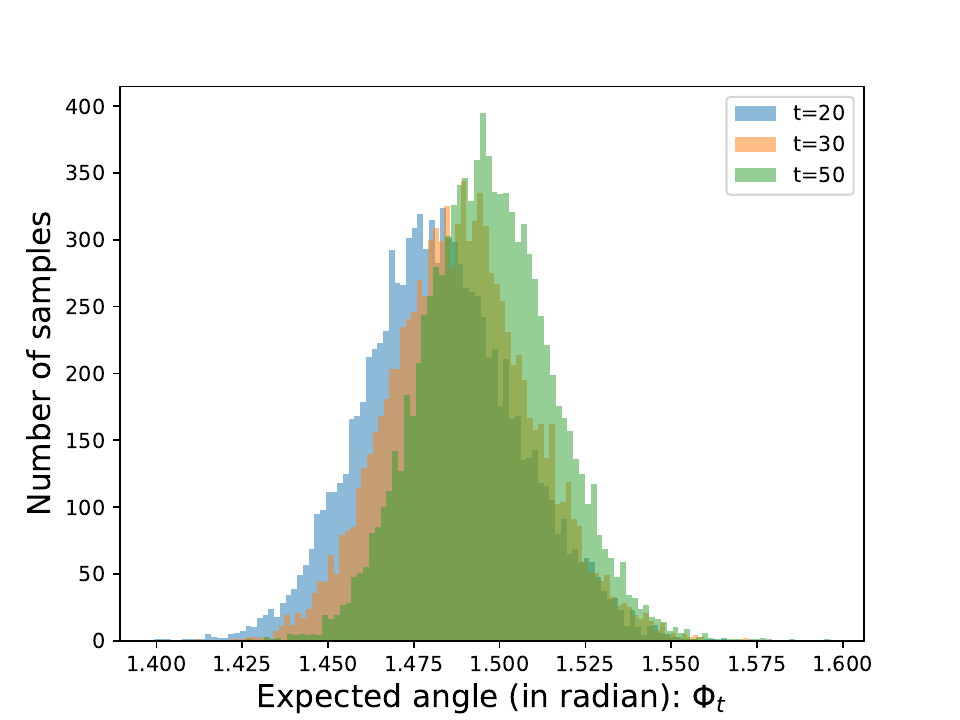}}
 
      \caption{histograms of $\Phi_t$ on the Reduced ImageNet and MNIST testing set. It shows an increment in the level of $\Phi_t$ along PGD-AT.}
\label{fig: angle_extra}
\end{figure}

\begin{figure}[!htbp]
    \centering
    \subfigure[Reduced ImageNet]{\includegraphics[width=0.32\textwidth]{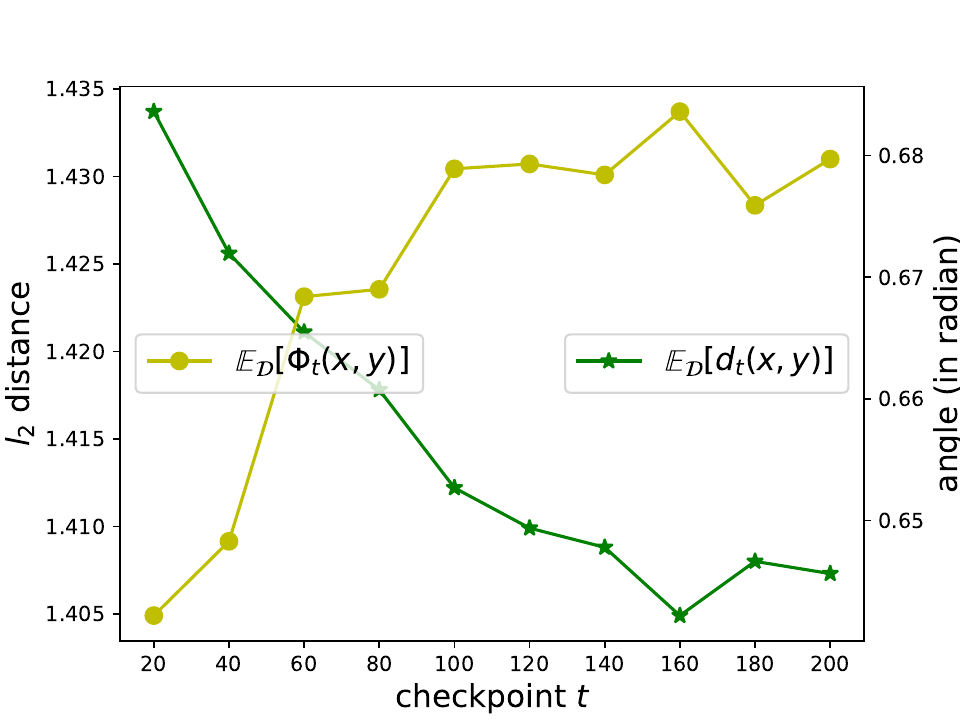}}
    \subfigure[MNIST]{\includegraphics[width=0.32\textwidth]{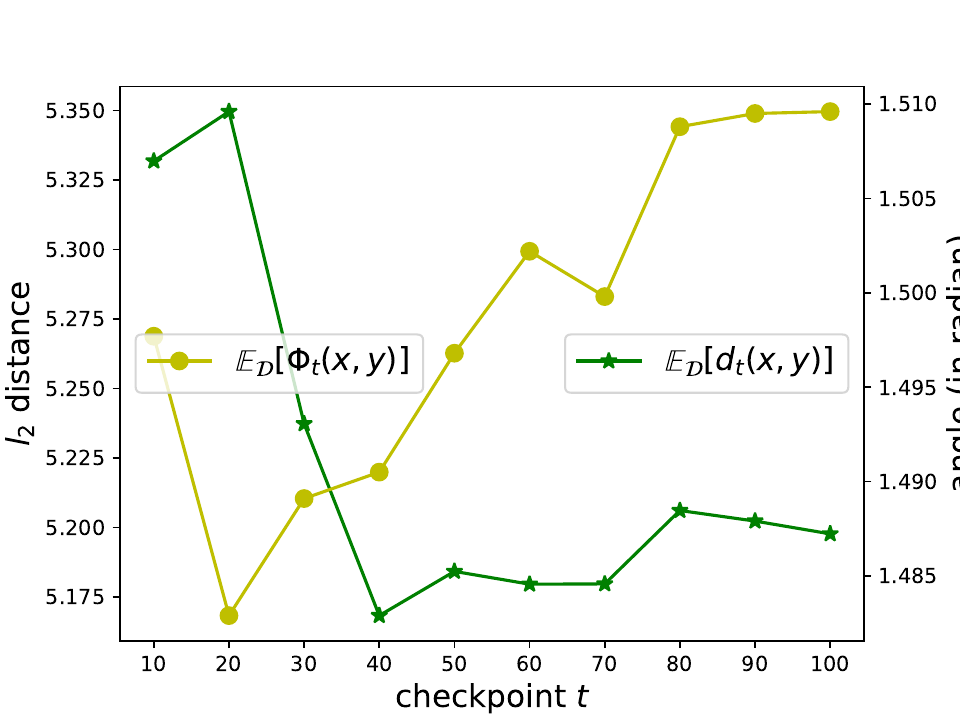}}
 
      \caption{The evolution of $\mathbb{E}_{\mathcal{D}}d_{t}(x,y)$ and $\mathbb{E}_{\mathcal{D}}\Phi_{t}(x,y)$ along PGD-AT evaluated on the testing set of the Reduced ImageNet and MNIST dataset.}
\label{fig: d2c_angle_extra}
\end{figure}

\begin{figure}[!htbp]
    \centering
    \subfigure[CIFAR-10 ($r=-0.971$)]{\includegraphics[width=0.32\textwidth]{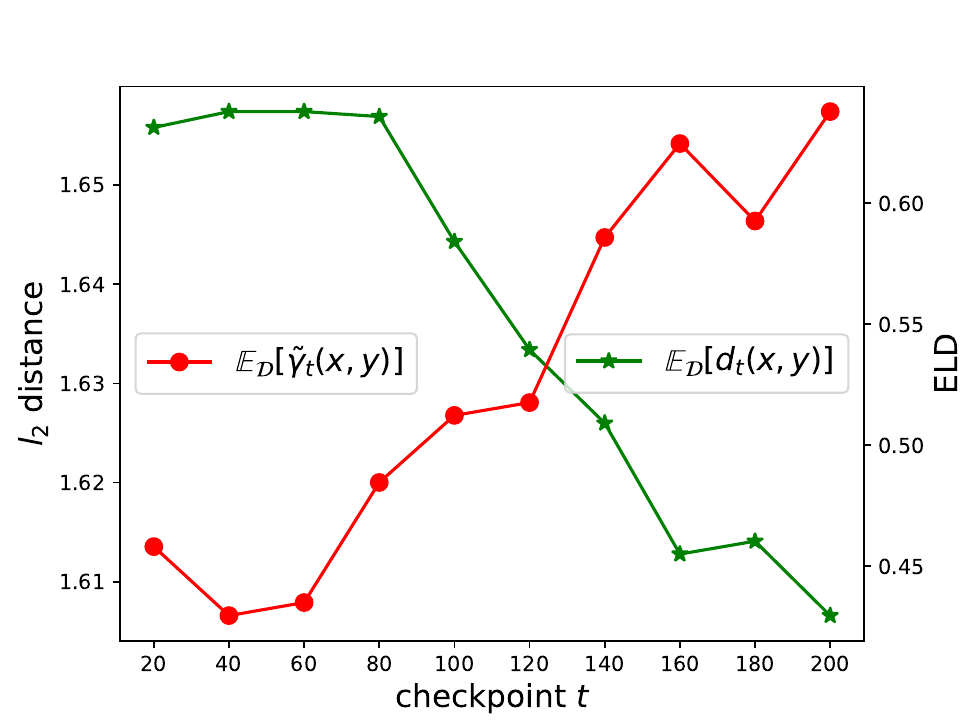}}
    \subfigure[MNIST ($r=0.917$)]{\includegraphics[width=0.32\textwidth]{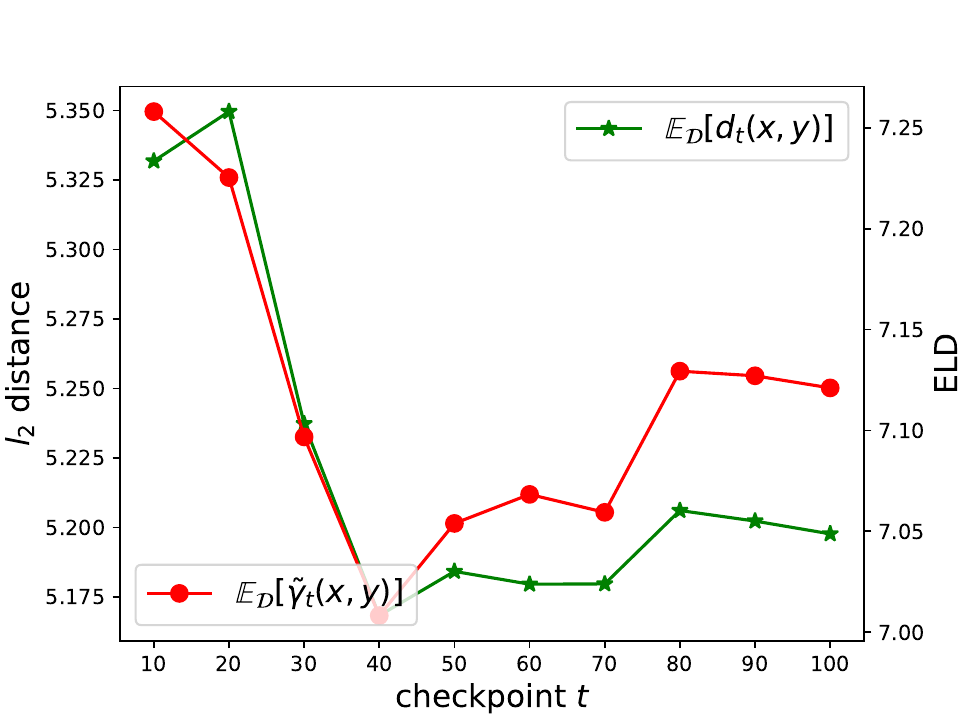}}
 
      \caption{We compare the evolution of $\mathbb{E}_{\mathcal{D}}d_{t}(x,y)$ (green curve) and $\mathbb{E}_{\mathcal{D}}\tilde{\gamma}_{t}(x,y)$ (red curve) on two different datasets to show that a negative correlation between these two quantities does not always exist.
      The value of $r$ in each sub-figures denotes the Pearson correlation coefficient between the two curves. It shows that on CIFAR-10 $\mathbb{E}_{\mathcal{D}}d_{t}(x,y)$ and $\mathbb{E}_{\mathcal{D}}\tilde{\gamma}_{t}(x,y)$ have a negative correlation whereas on MNIST a positive correlation appears.}
\label{fig: disper_d2c}
\end{figure}

\end{document}